\documentclass[sn-mathphys,Numbered]{sn-jnl}% Math and Physical Sciences Reference Style
\usepackage[utf8]{inputenc} 
\usepackage[T1]{fontenc}
\usepackage{graphicx}%
\usepackage{multirow}%
\usepackage{amsmath,amssymb,amsfonts}%
\usepackage{relsize}
\usepackage{xfrac}
\usepackage{nicefrac}
\usepackage{amsthm}%
\usepackage{mathrsfs}%
\usepackage[title]{appendix}%
\usepackage{xcolor}%
\usepackage{textcomp}%
\usepackage{manyfoot}%
\usepackage{booktabs}%
\usepackage{algorithm}%
\usepackage{algorithmicx}%
\usepackage{algpseudocode}%
\usepackage{listings}%
\usepackage{subcaption}
\usepackage[markup=underlined]{changes}

\theoremstyle{thmstyleone}%
\theoremstyle{thmstyletwo}%

\theoremstyle{thmstylethree}%

\begin{document}

\title[A Survey on Knowledge-Enhanced Multimodal Learning]{A Survey on Knowledge-Enhanced Multimodal Learning}


\author*[1]{\fnm{Maria} \sur{Lymperaiou}}\email{marialymp@islab.ntua.gr}

\author[1]{\fnm{Giorgos} \sur{Stamou}}\email{gstam@cs.ntua.gr}

\affil[1]{\orgdiv{Artificial Intelligence and Learning Systems Laboratory}, \orgname{School of Electrical and Computer Engineering,
National Technical University of Athens}, \orgaddress{\street{Heroon Polytechniou 9}, \city{Zografou}, \postcode{15780}, \state{Attica}, \country{Greece}}}


\abstract{Multimodal learning has been a field of increasing interest, aiming to combine various modalities in a single joint representation. Especially in the area of visiolinguistic (VL) learning multiple models and techniques have been developed, targeting a variety of tasks that involve images and text. VL models have reached unprecedented performances by extending the idea of Transformers, so that both modalities can learn from each other. Massive pre-training procedures enable VL models to acquire a certain level of real-world understanding, although many gaps can be identified: the limited comprehension of commonsense, factual, temporal and other everyday knowledge aspects questions the extendability of VL tasks. Knowledge graphs and other knowledge sources can fill those gaps by explicitly providing missing information, unlocking novel capabilities of VL models. At the same time, knowledge graphs enhance explainability, fairness and validity of decision making, issues of outermost importance for such complex implementations. The current survey aims to unify the fields of VL representation learning and knowledge graphs, and provides a taxonomy and analysis of knowledge-enhanced VL models.}

\keywords{Multimodal Learning, Vision-and-Language, Knowledge Graphs, Transformers}



\maketitle

\section{Introduction}\label{sec1}

Multimodal representation learning has been an area of machine learning that increasingly draws the attention of the AI research community. Combining information from different modalities, such as images, and text, allows more informative representations, as they provide complementary insights for the same instances. Several works focus on using both vision and language modalities, introducing tasks such as visual question answering \cite{vqa}, visual reasoning \cite{visual_reasoning, HOLZINGER2023100788}, visual commonsense reasoning \cite{vcr}, visual entailment \cite{entailment}, image captioning \cite{captioning}, image-text retrieval and inversely text-image retrieval \cite{retrieval}, referring expressions \cite{referring}, visual explanations \cite{explanations} and grounding \cite{grounding}, visual-language navigation \cite{navigation}, visual generation from text \cite{text-gen}, visual storytelling \cite{storytelling} and its inverse task of story visualization\cite{storygan}, and visual dialog \cite{dialog-gen}.

Some of the first attempts that combine vision and language face several limitations due to the restricted capacity of sequential models for language, such as recurrent neural networks (RNNs), LSTMs \cite{lstm} and GRUs \cite{gru}, which struggle to represent long textual sequences. The area of multimodal learning has faced significant advancements especially since the introduction of the Transformer framework \cite{transformer}. Several powerful transformer-based variants, such as BERT \cite{bert} and GPT-3 \cite{gpt3}, and more recently ChatGPT \cite{chatgpt} and GPT-4 \cite{gpt4} set the foundations for the surge of visiolinguistic (VL) transformers. The extension of single-modality pre-training requires the introduction of visual features, which allow to infer masked linguistic components and vice versa, enabling learning cross-modality relationships from aligned VL data. In the meantime, pre-training tasks applied independently on the visual or linguistic components permit intra-modality learning. Fine-tuning on task-specific VL datasets, addressing the tasks of text-image retrieval \cite{coco}, visual question answering \cite{coco_qa, vqa, gqa, visualgenome, Visual7W, cric}, visual reasoning \cite{nlvr}, visual commonsense reasoning \cite{vcr} and others, follows the pre-training stage. Models such as LXMERT \cite{tan2019lxmert}, VisualBERT \cite{li2019visualbert}, ViLBERT \cite{lu2019vilbert, lu202012in1}, VL-BERT \cite{su2020vlbert}, UNITER \cite{chen2020uniter}, OSCAR \cite{li2020oscar}, ViLT \cite{kim2021vilt}, CLIP \cite{clip}, SIMVLM \cite{simvlm}, FLIP \cite{flip} and many others have demonstrated state-of-the-art results in multiple VL tasks.

As for recent transformer-based approaches, despite pre-training on large amounts of aligned VL data, usually from Conceptual Captions, \cite{cc} COCO \cite{coco_qa} and Visual Genome \cite{visualgenome} datasets, the learned concepts remain limited and lack further explicit information regarding commonsense knowledge, abstract entities or real-world events. Relevant issues in the natural language processing field were addressed by leveraging knowledge graphs (KGs), thus resulting in knowledge-enhanced approaches for several tasks, such as Language Modeling \cite{know-nlp-2, know-nlp-6}, Natural Language Inference (NLI) \cite{know-nlp, know-nlp-8}, Language Generation \cite{know-nlp-1}, Dialog Generation \cite{know-nlp-4}, Entity Disambiguation \cite{know-nlp-5},  multilinguality \cite{know-nlp-7}, Contextualized Language Embeddings \cite{know-nlp-3}, and models such as KnowBert \cite{knowBERT}, E-BERT \cite{ebert}, ERICA \cite{erica}, ERNIE \cite{ernie}, ERNIE-NLI \cite{ernie-nli}, LUKE \cite{luke} and others. Therefore, the incorporation of large-scale knowledge graphs and ontologies can be also critical to the quality of multimodal representations and the success of relevant models on the various downstream tasks.

While previous surveys \cite{survey0, survey1, survey2, survey3, survey4, survey5, survey6, survey7} provide analysis and taxonomies over models, tasks and datasets regarding multimodal representations, they do not analyze knowledge-enhanced approaches. In contrast to those works, we focus on the integration and importance of external knowledge to VL models. Even though the current trends focus on transformer-based implementations, for the sake of completeness, other techniques that have contributed to the field of knowledge-enhanced VL (KVL) learning are also included. 
Overall, we target to bridge the gap between knowledge representation and multimodal deep learning: we provide a broad and comprehensive analysis of both fields, and consequently collect models that have served the various KVL tasks. Finally, we discuss current challenges and limitations of existing datasets and approaches, upon which we suggest potential future directions of this evolving field. 
To the best of our knowledge, there are no extended works covering the intersection of those two very fundamental fields of AI, which have demonstrated promising directions in many aspects of state-of-the-art research when combined.

The current survey consists of four main parts. The first part (Section \ref{sec:background}) covers the preliminaries of multimodal deep learning, analyzing trends, methods, models and tasks which set the basis for knowledge-enhanced VL (KVL) models. An analysis regarding graph structures and their representation follows in Section \ref{sec:graphs}. Section \ref{sec:knowledge} is dedicated to a taxonomy of knowledge senses and types, as well as the presentation of popular knowledge bases. Section \ref{sec:tasks} provides a taxonomy of KVL tasks  where the usage of external knowledge has been attempted, followed by dedicated Sections per corresponding task.  Finally, some needs and possible future directions regarding the usage of knowledge in multimodal learning are identified and analyzed in Section \ref{sec:future}.

\section{Background}
\label{sec:background}
Multimodal learning is a large and diverse field that involves a variety of data sources, architectural approaches and tasks. By focusing on VL tasks, which exploit text and image data, we can identify a variety of relevant applications. The nature of each task defines the chosen backbone architecture, upon which all consequent approaches are built. More specifically, VL tasks can be divided in \textbf{\textit{discriminative}} tasks, where the goal is either to provide a \textit{matching} between modalities or \textit{understanding} the one modality based on the other, and \textbf{\textit{generative}} tasks, which target either image generation (from text) or text generation (from image). 

\textbf{\textit{Discriminative}} VL tasks present a long line of research initially based on recurrent neural networks (RNNs) for text representation, with most contemporary approaches favoring the Transformer \cite{transformer} framework for its indisputable advantages. On the other hand, \textbf{\textit{generative}} tasks demonstrate an interesting variability in architectural approaches: while language generation tasks conditioned on image are addressed by architectures based on RNNs or Transformers, image generation tasks conditioned on text are mainly tackled by Generative Adversarial Networks (GANs) \cite{gan}, and more recently by Vision Transformers \cite{transformers-imgen} and Diffusion models \cite{diffusion}.

Knowledge-enhanced VL models (KVL models) usually step on existing approaches for VL representation and then employ various strategies to integrate knowledge.
One common first step between most VL approaches is the independent encoding of text \textit{T} and images \textit{I}, followed by the interaction of these encodings in order to acquire a joint representation. The choice of text encoding (RNN or Transformer) heavily influences the overall architecture of a VL model. On the other hand, image encoding adapts to the needs defined by text encoding, and variability in chosen image encoders serves particular improvements, such as performance boosting or reduction of trainable parameters. 

There is a variety of ways to incorporate \textit{knowledge} \textit{K}, with many approaches favoring \textit{external knowledge sources} in the form of widely used knowledge graphs (KGs). In this case, 
a representation based on graph neural networks (GNNs) is the most popular approach towards providing an appropriate encoding for \textit{K}. However, other knowledge types can be integrated as well, either in the form of knowledge stored in neural network weights (implicit knowledge) or linguistic knowledge from the web, embedded with a text encoder. Enhancements to VL model performance can be also realized via \textit{self-acquired} \textit{(internal) knowledge}: without leveraging external knowledge sources, automatic extraction of specific characteristics from either modality boosts and guides learning, improving knowledge-free baselines.
Knowledge \textit{K} can be fused either in early stages together with text $T\in \mathcal{W}$ and image $I\in \mathcal{V}$ instances resulting in a KVL representation, or alternatively in later stages and independently of the VL stream, refining and correcting the predictions of the VL model.

KVL models can either target one task at a time (\textbf{\textit{single-task models}}), or multiple tasks simultaneously (\textbf{\textit{multi-task models}}). Single-task models present a large variety of architectural implementations so far, while multi-task models are exclusively built upon multimodal transformers, as they heavily rely on the pre-training fine-tuning scheme.  \textbf{\textit{Discriminative}} tasks are tackled by both single- and multi-task models. On the other hand, \textbf{\textit{generative}} tasks can be only handled by single-task models, as they are harder by nature \footnote{Generative tasks involve generating entire data samples, which requires modeling the joint distribution of input and output variables. This often involves dealing with high-dimensional and complex output spaces, making it inherently more challenging than discriminative tasks, which only need to learn the conditional distribution of the output (label) given the input. Even by considering the two task categories from a human's perspective, it is fairly easier to recognize or understand data given a finite set of samples (e.g. detect and classify objects in an image) rather than create the data ourselves (e.g. accurately paint objects to match those existing in a given training image dataset -especially in cases we have never encountered the depicted object classes, as in the case of a model learning from scratch-).
Moreover, ambiguity in outcome evaluation (involving possible variations and uncertainty) and consequently in related loss functions makes the development of generative models challenging, as features and directions towards outcome improvement are not straightforward.}. The evaluation of the overall model performance is realized as per task, based on appropriate task-specific evaluation metrics.

\begin{figure}[h!]
    \centering
    \begin{subfigure}[b]{0.97\textwidth}
        \includegraphics[width=\textwidth]{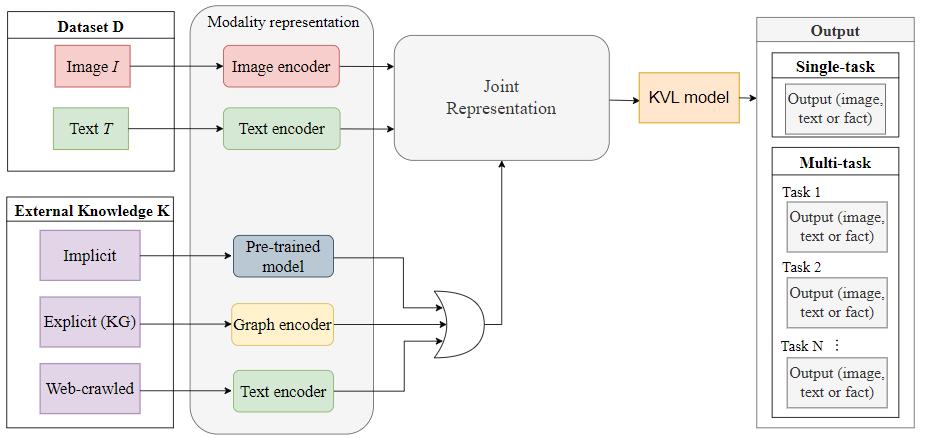}
        \caption{KVL pipeline with early knowledge fusion}
        \label{fig:sub1}
    \end{subfigure}
    \begin{subfigure}[b]{0.97\textwidth}
        \includegraphics[width=\textwidth]{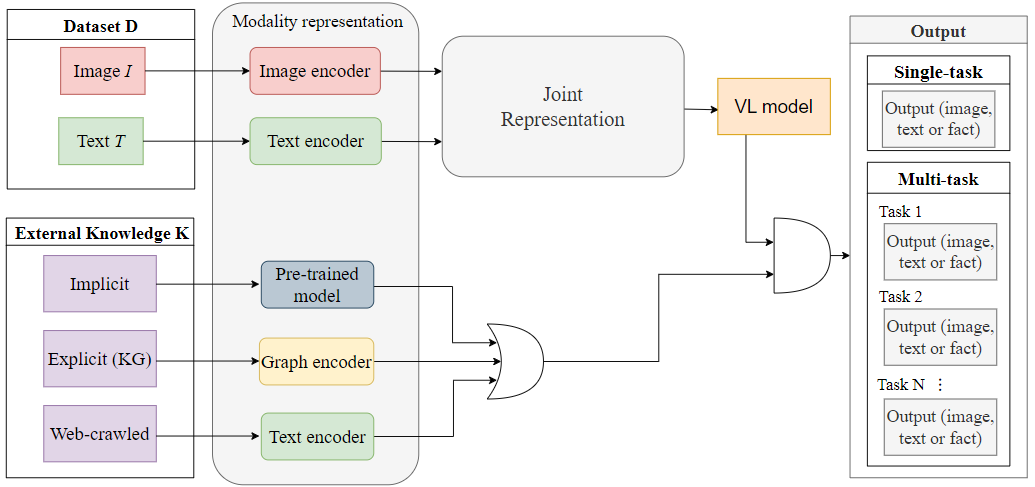}
        \caption{KVL pipeline with late knowledge fusion}
        \label{fig:sub2}
    \end{subfigure}
    \caption{A generic overview of KVL implementations. Knowledge can be fused early in the pipeline, contributing to a KVL representation, or later on, modifying the outcome of a VL model.}
    \label{fig:overview}
\end{figure}

A broad overview of KVL models is provided in Figure \ref{fig:overview}. The joint representation module is shown as a black box, as most architectural variations analyzed in following Sections are happening within this stage.

\section{Multimodal representation learning}
\label{sec:multimodal}
The core of multimodal deep learning revolves around the ways the various modalities are represented independently and interact with each other. Especially for text and images, several representation techniques have been developed throughout the years, due to the advancements of image classification \cite{resnet, vggnet} and object detection \cite{fast_rcnn, faster_rcnn} models for vision, as well as distributed language representations \cite{word2vec, glove}, recurrent architectures \cite{lstm, bilstm, gru}, and attention-based models \cite{transformer, bert, roberta, gpt3} for text. Apart from vision and language, more modalities can potentially contribute in a joint representation, such as speech, music, graphs and others.

\subsection{Text representation}
\label{sec:text}
Text representation offers several variations over implementations, imposing significant influences towards multimodal learning in total. The two major categories of language embeddings include \textbf{recurrent architectures (RNN/LSTM/GRU)} and \textbf{Transformers}. 
The architectural choices for language representation guide the taxonomy of models per KVL task, due to the diversity they impose in the resulting implementations. 

\paragraph{Distributed word representations} 
Traditional and widely used representations such as Word2Vec \cite{word2vec} have contributed in several components of the various VL tasks, often used as initializations for other methods. 
Doc2vec \cite{doc2vec} extends word2vec, achieving a vector representation of a group of words. GloVe \cite{glove}, a distributed word representation model, is trained on a global word-word co-occurrence frequency matrix and successfully captures both local and global statistics.  Fast-text \cite{fasttext} represents each word with a bag of character n-grams in order to capture the internal structure of words. This approach can effectively utilize the morphology of words, thus it is able to represent rare formations occurring in morphologically rich languages.  Word semantics cannot be successfully captured by static word representation methods, so contextualization is needed especially in cases of polysemy, an issue tackled in CoVe \cite{cove}. ELMo \cite{elmo} is another deep word contextualization model which leverages character level information to form robust representations based on morphological information.  
Universal Language Model Fine-tuning (ULMFiT) \cite{ulmfit} achieves robust inductive transfer learning on a variety of downstream NLP tasks, where only fine-tuning is required.

\paragraph{Recurrent neural networks (RNNs)}
The basic idea behind RNNs is the sequential processing of elements belonging to a finite sequence $(x_1, x_2, ..., x_T)$, one $x_t$ at a time, while retaining context information from the previous elements in the form of the previous node's hidden state $h_{t-1}$. Both $x_t$ and $h_{t-1}$ contribute to the calculation of the current hidden state $h_{t}$, which consequently participates in defining the current output $y_{t}$. Feed forward neural networks implement each time step as a layer, with shared weights across layers. Backpropagation through time (BPTT) is utilized to train recurrent neural networks \cite{rnn-review}.  Other recurrent neural architectures such as Long
Short-Term Memory (LSTM) \cite{lstm} dominated the field of language embeddings in many approaches, with more refined variants such as bidirectional LSTM (BiLSTM) \cite{bilstm}, Gated Recurrent Unit (GRU) \cite{gru} and bidirectional GRUs (BiGRUs) following in later works. 

This sequential nature of RNNs fits naturally to language processing, inspiring several implementations in the NLP domain, as well as in multimodal tasks. To this end, earlier works in VL architectures heavily rely on distributed word embeddings together with sequential models for language representation. However, limitations tied with sequential processing such as vanishing gradients and inability of parallel processing direct research interest towards novel approaches, such as the Transformer.

\paragraph{Language transformers}
The introduction of attention mechanisms and especially the Transformer \cite{transformer} opened a whole new world of possibilities for language embeddings and several downstream linguistic tasks. 
The Transformer relies on encoder-decoder structure and utilizes an attention mechanism to construct dependencies between input and output data. 
%The number of encoder and decoder layers is a hyperparameter, and N=6 is chosen in the original paper.
The layers of the encoder and the decoder are stacked the one upon the other, containing sub-layers with Multi-Head self-attention and position-wise fully connected feed-forward layers. Residual connections followed by layer normalization are used between the sub-layers of the encoder. The decoder has an additional encoder-decoder multi-head attention sub-layer that helps focusing on the appropriate parts of the encoded input sequence. Moreover, the decoder's self-attention modules are modified so that they force the prediction at a certain position to be based on the known predictions of previous positions exclusively. Transformer architectures prove that there is no need for convolutions or recurrent units to achieve state-of-the-art performance in linguistic tasks. Currently, most state-of-the-art VL architectures utilize attention mechanisms within their implementation.

Transformer models for NLP tasks consist of a \textbf{\textit{pre-training}} stage on a large corpus of data, followed by \textbf{\textit{fine-tuning}} on certain downstream tasks. Language transformers can be divided in two major categories: \textbf{\textit{autoregressive (AR)}} and \textbf{\textit{autoencoding (AE)}}, depending on whether pre-training is performed in a \textit{unidirectional} or a \textit{bidirectional} way. AR language models attempt to estimate the probability distribution of a text corpus, while AE models learn to reconstruct manipulated inputs with the help of surrounding information \cite{xlnet}.

BERT \cite{bert} is a popular bidirectional transformer-based language model, able to handle a variety of natural language processing tasks by just fine-tuning one additional output layer. It uses masked language modeling (MLM), randomly hiding some input tokens in order to be inferred from the surrounding words. The \textbf{\textit{pre-training}} stage is based on unlabeled data, which enable parameter learning. Those parameters are then fine-tuned with labelled data corresponding to some certain tasks. 
RoBERTa \cite{roberta} offers an optimized extension to BERT, suggesting that longer training, more data and larger batch size, as well as training on longer sequences, dynamically altering the masking patterns and removing the next sentence prediction loss are factors that contribute to advanced performance of the original BERT model. XLNet \cite{xlnet} combines AR and AE language modelling in a single approach. It introduces the pre-training objective of permuted language modelling, which attempts to collect information from permutations of the factorization order of text tokens with respect to AR likelihood estimation, practically inducing bidirectional capabilities to the learning process. 

Generative NLP models have demonstrated impressive results in recent literature. GPT-2 \cite{gpt2} is pre-trained on a very large dataset (40GB of text data from over 8 million documents) and utilizes 1.5 billion parameters in order to construct a powerful language model. 
%Scaling up in unprecedentent number of parameters by the time of its development, GPT-2 indicated that the usage of more parameters results in higher performance in a log-linear fashion on a variety of downstream tasks, while zero-shot behavior are enabled.
GPT-3 \cite{gpt3} is an AR language model of 175 billion parameters which achieves zero-shot, one-shot and few-shot capabilities. It is able of synthesizing results in a human-like level, so that writing articles or code in some programming language, learning and reusing new words, unscramble words and other tasks can be realized indistinguishably to humans.
T5 \cite{t5} introduces a unified format where both inputs and outputs are text. Without changing the model architecture, the hyperparameters or the loss function, T5 generates text to address tasks such as question answering, text summarization, machine translation, as well as classification tasks. 
BART \cite{bart} utilizes a bidirectional encoder, enabling attention from both left and right directions, and an autoregressive (unidirectional) decoder, which allows attending to past tokens only. It is able to handle generative tasks such as question-answering, text summarization, conditional text generation, mask filling, and also classification tasks. ChatGPT \cite{chatgpt} (based on the GPT-3.5 architecture) represents a significant advancement in generative NLP compared to its predecessors, thanks to its ability to generate coherent and contextually relevant text across various domains. With its large model size (175 billion parameters), improved training techniques, and enhanced contextual understanding, ChatGPT sets a new standard for natural language generation, enabling more engaging and human-like interactions. Overall, the surge of Large Language Models (LLM) has enabled a plethora of architectures for language generation, including GPT-4 \cite{gpt4}, LLaMA/LLaMa 2 \cite{touvron2023llama, touvron2023llama2}, Vicuna \cite{vicuna2023}, Falcon \cite{almazrouei2023falcon}, Mistral \cite{jiang2023mistral}, Mixtral, \cite{jiang2024mixtral}, Gemini \cite{team2023gemini} among others.

As for text encodings for VL models, BERT \cite{bert} has become a gold standard for several transformer-based approaches, while fewer implementations utilize variants such as RoBERTa \cite{roberta}. GPT2 \cite{gpt2}, T5 \cite{t5} and BART \cite{bart} have also served as language encoders.

\subsection{Visual representation}
\label{sec:vision}
There is much less diversity in the representation of the visual modality compared to text encoding. Most works rely on widespread \textit{convolutional} architectures without significant variations, and only recently some works attempted encoders based on \textit{image transformers}, which however do not enforce architectural modifications.

\paragraph{Convolutional Neural Networks (CNNs)}
Representation of images can involve object level or image level features, depending on the granularity of information that needs to be imbued in the representation. A global image representation can be achieved by employing widely used image classification models as feature extractors. Many works rely on CNN based classifiers such as VGG \cite{vggnet} and ResNet \cite{resnet}, while others prefer more fine-grained local representations supported by object detectors, such as Fast-RCNN \cite{fast_rcnn} and Faster-RCNN \cite{faster_rcnn}. Speed in object detection was promoted in the YOLO model family \cite{yolo}, which parses the whole image only once; consequent YOLO variants \cite{wang2022yolov7} increase model size to ensure higher accuracy.
The fixed pre-trained models for object feature extraction somehow limit the expressivity of VL transformers, while being slow. Solutions to this limitation include the usage of grid features as visual tokens \cite{pixelbert}, or discretized grid features \cite{soho}. 

\paragraph{Image Transformers} Recent advancements in Vision Transformers as an extension of the aforementioned Language Transformers \cite{transformer} have influenced the field of image representation. ViT \cite{vit} suggests patch-like parsing of images for feature extraction, resulting in more powerful image representations. 
Swin Transformer \cite{Swin} is a more efficient implementation due to the usage of self-attention in local image patches contrary to global self-attention of other approaches, which results in quadratic computation complexity compared to image size. Swin Transformer achieves linear complexity by hierarchically merging larger and larger image patches across layers, with self-attention acting only within each patch. Similar to NLP Transformers scaling capabilities, Swin Transformer V2 of 3 billion trainable parameters serves as one of the largest dense vision models so far \cite{swinv2}. Recently, more transformer-based object detectors \cite{xu2021endtoend, duan2022centernet, chen2023vision, zong2023detrs} have demonstrated state-of-the-art results.

\subsection{Sequential models for VL tasks}
Even though Transformer-based approaches have dominated the field of multimodal learning, sequential models have offered a variety of interesting solutions in several tasks, and still serve as the way to go in some of them. Most sequential-based techniques address the vision-language co-operation through encoding-decoding schemes which utilize a CNN for images \textit{I} and an RNN/LSTM/GRU structure for text \textit{T}. There are different ways for visual and language embeddings to interact, depending on the downstream task: for example, a CNN-based image encoding can be fed as a conditioning to an RNN/LSTM/GRU decoder structure for tasks requiring text generation from image. Alternatively, input text can be embedded using an RNN/LSTM/GRU encoder, and then a feed-forward neural network can learn the correlations between text embeddings and CNN-based image embeddings. This variant can also serve text-image matching tasks. Sequential structures for VL learning remain rather popular in tasks that require language generation, especially in underexplored ones where more refined architectures have not been attempted yet.

\begin{figure}[htp!]
    \centering
    \includegraphics[width=\textwidth]{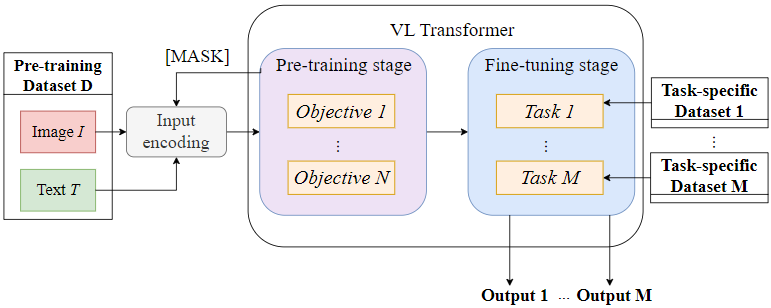}
    \caption{The overall workflow of a VL transformer.}
    \label{fig:overall}
\end{figure}

\subsection{Multimodal Transformers}
Multimodal transformers have revolutionized the field of multimodal learning, with almost any state-of-the-art model built upon them. There are some certain steps followed from receiving the input data until the final result on multimodal tasks, as presented in Figure \ref{fig:overall}. Initially, given a multimodal dataset, for example a VL dataset $D$ comprised of image-text pairs $(I, T)$ which consist of visual features $v$ and textual features $w$ respectively, an appropriate encoding scheme, such as the one described in Sections \ref{sec:text} for \textit{T} and \ref{sec:vision} for \textit{I} should be decided. 
The input embedding contains a tokenized text representation, an image encoding and other special embeddings.

All transformer-based architectures, either targeting vision, language or both consist of the \textbf{\textit{pre-training}} and
\textbf{\textit{fine-tuning}} stages.  For this reason, a multimodal encoding module is designed to receive the embedded input, enabling the interaction between modalities by jointly learning complementary information with the help of well-designed cross-modal pre-training tasks. Finally, a fine-tuning stage adapts the pre-trained model to the downstream task by training on a smaller labelled dataset.

\begin{figure}[h!]
    \centering
    \includegraphics[width=0.98\textwidth]{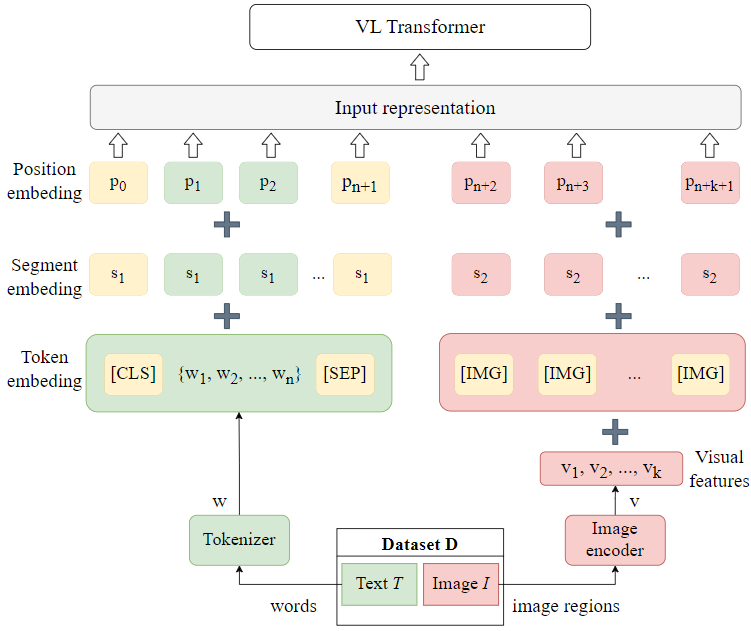}
    \caption{A general outline of input tokens and embeddings.}
    \label{fig:input}
\end{figure}

\subsubsection{Special input tokens and embeddings}
Special tokens need to be appended to the input before entering the VL transformer model in order to discern between different modalities, as well as the start and sometimes the end of the sequence. An input embedding is formed by combining text representation, visual representation, special tokens and other embedding information to guide training. Despite different VL transformers following slightly different strategies regarding input representation, the main constituents are analyzed below. Figure \ref{fig:input} provides a general overview of the input tokens and embeddings.

The \textbf{input token} denoted as [CLS] is a \textit{special classification} token that defines the start of the input sequence. Linguistic information from text \textit{T} is appended after the [CLS] token. Usually WordPiece \cite{wordpiece} tokenizer, a sub-word based tokenizer framework, transforms words into tokens. Afterwards, a numerical format of tokens is obtained by assigning a unique embedding per subword token so that it can be further processed. The \textbf{text embedding} will then be  $w = {w_1, w_2, ..., w_n} \in \mathbb{R}^d$, where $n$ corresponds to the textual sequence length and $d$ is the embedding dimension. A \textbf{segment token} [SEP] is appended after $w$ to separate different modalities. Following the separator [SEP], a \textbf{visual token} [IMG] indicates the visual modality \textit{I}. There are different usages of [IMG], either marking the start of visual features $v$, or multiple [IMG] instances acting as masks that locate the existence of visual elements within the \textbf{token embedding}. Some architectures may even discard [IMG] tokens. A flattened visual feature vector is extracted as a result of the chosen visual encoder, denoted as $v = {v_1, v_2, ..., v_k} \in \mathbb{R}^d$, where $k$ is the number of visual features and $d$ the embedding dimension.  The \textbf{end token} [END] marks the end of the sequence. Many VL transformers ommit [END] token.

\textbf{Segment embeddings} attribute the source of each input element by assigning a unique label to each of them. For example, input textual and visual features would be assigned with a different \textbf{segment embedding} $s_w$ and $s_v$ respectively, as they come from different modalities. Therefore, sequences $s_w = s_{w_1}, s_{w_2}, ..., s_{w_n}$ are going to be summed with the \textbf{text embeddings} and similarly, $s_v = s_{v_1}, s_{v_2}, ..., s_{v_k}$ will be summed with \textbf{visual embeddings}.
Alternatively, if the input contains two sentences (such as question and answer pairs), each of those will be aligned with different segment labels. \textbf{Position embeddings} $p_i$ denote the positioning of tokens in a sequence $\{w_1, ..., w_i, ..., w_n\}$, as originally used in BERT \cite{bert}. 

The summation of \textbf{token embeddings},\textbf{ segment embeddings}, \textbf{position embeddings} and \textbf{visual embeddings} can act as starting point towards the \textbf{input representation}, followed by an appropriate transformer structure. The \textbf{input embedding} generally follows the $\{[CLS], \hat{w}_1, \hat{w}_2,$ ... $, \hat{w}_n, [SEP], \hat{v}_1, \hat{v}_2, $ ...$, \hat{v}_k, [END]\}$ format, where $\hat{w}$ and $\hat{v}$ are the final \textbf{text} and \textbf{visual embeddings} respectively, after the summation of $w$ and $v$ with the \textbf{segment} and/or \textbf{position embeddings}.

\subsubsection{Vision and Language joint encoding}
Encoded modalities need to be projected to the same vector space and interact in order to achieve a meaningful joint representation. A concise separation of cross-modality encoding is provided in \citet{survey7}, where two main categories of encoding schemes are identified: \textbf{\textit{fusion encoder}} and \textbf{\textit{dual encoder}}. Those two approaches can be even combined. Fusion encoder concerns an abundance of VL transformers, and can be further divided in \textbf{\textit{single-stream encoding}} and \textbf{\textit{double-stream encoding}}.

\paragraph{Double-stream fusion encoder} Double-stream fusion encoder utilizes two separate transformer modules to process images and text respectively. First VL approaches such as ViLBERT \cite{lu2019vilbert} and LXMERT \cite{tan2019lxmert} fall into this category, where they naturally extend BERT to also process images. Specifically, ViLBERT decomposes images in non-overlapping patches, similar to how tokens serve as inputs in the case of BERT. Text is tokenized and fed together with positional embeddings to its transformer stream. A co-attention module enables interaction and alignment between modalities given their intermediate representations. An extension of ViLBERT to 12 downstream tasks was presented in \citet{lu202012in1}. LXMERT \cite{tan2019lxmert} refines the cross-modal part to achieve advanced performance in downstream tasks. More recent models employing a \textbf{\textit{double-stream fusion encoder}} are ALBEF \cite{albef}, Visual Parsing \cite{visual-parsing} and WenLan \cite{wenlan}.
In general, \textbf{\textit{double-stream encoders}} demand training of two transformer models (one for each stream), which is computationally inefficient. Therefore, succeeding approaches focus on \textbf{\textit{single-stream encoders}}.

\paragraph{Single-stream fusion encoder} Single-stream fusion encoder concerns the majority of state-of-the-art VL models. A single transformer network, usually stepping on the BERT \cite{bert} backbone, is used to process images and text representations simultaneously, where alignments are discovered via self-attention. Segment tokens to separate modalities, together with position tokens to indicate aligned token pairs are added to the input data before they are concatenated and fed into the encoder. VisualBERT \cite{li2019visualbert}, VL-BERT \cite{su2020vlbert}, PixelBERT \cite{pixelbert}, InterBERT \cite{lin2021interbert}, VLP \cite{vlp}, Unified-VLP \cite{u-vlp}, B2T2 \cite{b2t2}, UNITER \cite{chen2020uniter}, XGPT \cite{xgpt}, ViLT \cite{kim2021vilt}, VL-T5 \cite{vlt5}, SOHO \cite{soho}, SimVLM \cite{simvlm} belong to the \textbf{\textit{single-stream encoder}} category. Some of the latest models even present zero-shot learning capabilities, enabling more out-of-the-box capabilities in VL tasks.

\paragraph{Dual encoder}  The Dual encoder is favored by a small family of recent VL models which exploit contrastive learning to provide image-text representations. Separate encoders embed each modality independently, and their representations are projected on the same vector space with the goal of learning similarity and dissimilarity properties. This is why contrastive learning naturally fits: paired image-text samples are trained to stay close together, while being apart from the rest. CLIP \cite{clip} implements this strategy utilizing more than 400 million image-text pairs for training, achieving even zero-shot capabilities in retrieving previously unseen matches. ALIGN \cite{align} follows the same recipe using over one billion image-text pairs, demonstrating that large-scale uncurated data can compensate for the presence of noise. FLORENCE \cite{florence} exploits both image-to-language contrastive loss and its reverse, language-to-image contrastive loss, actually forming a bi-directional contrastive loss applied on image-label-description triplets.

\paragraph{Fusion and dual encoder} A couple of models leverage the benefits of both encoding approaches \cite{survey7}. FLAVA \cite{flava} is a holistic approach that utilizes a \textbf{\textit{dual encoder}} to integrate visual and text representations into a multimodal one, providing unimodal and multimodal reasoning capabilities in the same model. Additionally, VLMo \cite{wang2021vlmo} offers a flexible format by providing a \textbf{\textit{fusion encoder}} for classification task, and a \textbf{\textit{dual encoder}} for retrieval tasks.

\subsubsection{Self-supervised pre-training}
\label{sec:pretraining_datasets}
\paragraph{VL pre-training datasets}
A VL model can be pre-trained on \textit{unimodal} (text and/or visual data independently) and \textit{multimodal} data (paired image-text data). \textit{Unimodal} data can be unlabelled, noisy and abundant, such as sets of documents or images, while \textit{multimodal} data are labelled and cleaner, so that each modality presents another point of view for the same instance.
Large-scale datasets containing visual and linguistic information are necessary for pre-training. Pre-training aspires to instill a generic understanding of the visual world, natural language and their in between interactions. Widely used datasets can either be \textit{in-domain}, meaning that their data distribution is very close to task-specific datasets used in the fine-tuning stage, or \textit{out-of-domain}, containing less similar data to the downstream tasks, but usually being much larger in size. Most VL transformers leverage a corpus consisting of COCO \cite{coco}, Visual Genome \cite{visualgenome}, SBU \cite{sbu} and Conceptual Captions (CC) \cite{cc} for pre-training, with fewer models either excluding some of those, or adding more datasets to this corpus.

\textbf{COCO} \cite{coco} consists of around 106K images/533K captions in the train split, and 25K images/5K captions in the test split, with 5 sentences provided per caption from different annotators. Those sentences are designed to provide an overall (global) understanding of their corresponding images, which represent scenes of multiple objects and their in-between relationships. As most VL tasks are built atop COCO, it is considered to be an \textit{in-domain} dataset. Because of the usage of COCO for downstream tasks, subsets of COCO used in pre-training and fine-tuning must be mutually exclusive to avoid any data leakage.

\textbf{Visual Genome (VG)} \cite{visualgenome} is another \textit{in-domain} dataset, from which several VL tasks emerge. It contains more than 100K images of complex scenes, providing multiple annotations regarding per image scene graphs, objects (3.8M instances), relationships (2.3 M), attributes (2.8 M), visual questions and answers (1.7 M), region descriptions (5.4 M) and region scene graphs (3,7 M). Region descriptions are captions grounded to image regions, acting as dense captions per image. As VG images have a partial overlap with COCO images, any COCO image used in downstream tasks should be excluded from the VG during pre-training.

\textbf{SBU Captions} \cite{sbu} consists of 990K images/990K captions in the train split, and 10K images/10K captions in the test spit, with each image corresponding to one caption. It is an \textit{out-of-domain} dataset, larger in size than COCO and VG. 

\textbf{Conceptual Captions (CC)} \cite{cc} is the largest \textit{out-of-domain} dataset used for pre-training, consisting of more than 3 million images for training and 14K for validation with 1 caption each.

\paragraph{Pre-training objectives}
\textit{Uni-modal objectives} concerning either \textit{T} (\textit{language objectives}) or \textit{I} (\textit{vision objectives}) at a time, as well as \textit{cross-modal objectives}, which take into  account both modalities at once are used for pre-training. Such objectives teach the models to infer missing information by understanding their surroundings. Self-supervised learning is the most prevalent technique, enabling learning with the help of a corrupted part of the input or adversarially matched pairs in a contrastive fashion.

\textit{Language objectives} are designed to implicitly learn linguistic rules and patters so that a model pre-trained on them acquires some 'understanding' of natural language. Some language modelling objectives are analyzed below. \\
\textbf{\textit{Masked Language Modeling (MLM)}} is a pre-training objective introduced in BERT \cite{bert}. Tokens in the input sentence are replaced at random with a special [MASK] token. The model needs to uncover the actual token by learning the context from its surroundings, thus permitting contextualized representations. The default probability of masking out a token is 0.15. The MLM objective function is bidirectional, which means that a token can be predicted from either its right ones, or its left ones. \textbf{\textit{Prefix Language Modeling (PrefixLM)}} \cite{simvlm} differs from standard MLM such that it enables bi-directional attention on the prefix sequence.
\textbf{\textit{Next Sequence Prediction (NSP)}} refers to the objective that given a pair of sentences, tasks the model to predict whether they could serve as consecutive sentences in a corpus or not.\\

\textit{Vision objectives} apply similar ideas on the visual modality. Due to the more high-dimensional nature of images in comparison to text, the masking tasks are challenging in design. 

\textbf{\textit{Masked Region Modeling (MRM)}} applies zeros over image regions, asking the model to infer missing parts. Semantic and pixel level information can be retrieved, naturally leading to two sub-tasks:
\begin{itemize}
    \item \textbf{\textit{Classification}} of regions can be used to obtain a discrete signal corresponding to \textit{semantic} entities. 
    \item \textbf{\textit{Regression}} of region features provides a more continuous, \textit{pixel-level} understanding of missing parts.
\end{itemize}
\textbf{\textit{Random Pixel Sampling (RPS)}} \cite{pixelbert} tackles overfitting similarly to the Dropout mechanism \cite{dropout}: in each iteration, only a subset of pixels is chosen to be inserted in the transformer network. The corrupted image that the model receives each time enables more robust representations, as relying on \textit{semantics} instead of single \textit{pixels} is encouraged. \\

\textit{Cross-modal objectives} jointly leverage information from both text and image modalities for learning. The design of \textit{cross-modal} pre-training tasks is more challenging compared to their \textit{unimodal} counterparts, as it is required to ensure that the model does not depend learning on a single modality exclusively
\cite{chen2020uniter, simvlm, survey7}. 

\textbf{\textit{Unidirectional (seq2seq) Language Modeling (seq2seqLM)}} is an autoregressive variant of MLM, which means that a masked token can only attend to previous tokens and not to future ones. Seq2seqLM attempts to directly maximize the likelihood of a text- image pair $(T, I) = x \in (\mathcal{W}, \mathcal{V})$ from a dataset $\mathcal{D}$:
\begin{equation*}
    \mathcal{L}_{LM} = -\mathbb{E}_{(\mathcal{W}, \mathcal{V}) \in D}\log P_{\theta}(x))  \\ 
\end{equation*} 
%\hfill \break
\textbf{\textit{Masked Language Modeling with Image (MLMI)}} attempts to recover the corrupted text tokens by also consulting the image apart from the linguistic part exclusively. Inferring the masked token is achieved by minimizing the negative log-likelihood, where $w_m$ are the masked tokens, $w_{\setminus m}$ are the unmasked ones, and $(\mathcal{W}, \mathcal{V}) \in \mathcal{D}$: 
\begin{equation*}
    \mathcal{L}_{MLM} = -\mathbb{E}_{(\mathcal{W}, \mathcal{V}) \in D}\log P_{\theta}(w_m|w_{\setminus m}, \mathcal{V}) \\    
\end{equation*}
\hfill \break
\textbf{\textit{Whole Word Masking (WWM)}} \cite{kim2021vilt} is an extension of MLMI which masks out entire words rather than subword tokens. Therefore, the model is encouraged to consult the visual part to infer the corrupted word instead of guessing the missing part from its unmasked constituents. This procedure enforces a more difficult task to the transformer, achieving better cross-modal alignment.  \\
\textbf{\textit{Masked Region Modelling with Language (MRCL)}} is the complementary task of MLMI. Instead of text tokens, image region features are masked out with a probability (by default 0.15). Masking does not include a special token for the visual modality; it is implemented by filling the corresponding image regions with zeros. The model is tasked to reconstruct visual regions $v_{\setminus m}$ based on information provided from the unmasked features $v_{m}$ and the textual modality $\mathcal{W}$, targeting to optimize the objective: 

\begin{equation*}
    \mathcal{L}_{MRM} = -\mathbb{E}_{(\mathcal{W}, \mathcal{V}) \in D}f_{\theta}(v_m|v_{\setminus m}, \mathcal{W}) 
\end{equation*} 
\hfill \break
Reconstruction of visual features can yield two tasks, offering different insights to the high-dimensional problem of feature prediction:
\begin{itemize}
    \item \textbf{Masked Region Feature Regression (MRFR)} refers to producing features of the same dimensionality as the visual region. This is achieved by applying $L_2$ regression between the predicted and the ground truth visual vector to minimize their distance. Considering the transformer prediction $FC(h_{v_i})$, acquired after passing the output of the masked region $h_{v_i}$ through a fully connected (FC) layer, and the region feature $\hat{E}_v(v_i)$ of region $v_i$, the minimization of $L_2$ can be written as:
    \begin{center}
           $\mathcal{L}_{MRFR}$ = -$\mathbb{E}_{(\mathcal{W}, \mathcal{V}) \in D}\mathlarger{\mathlarger{\sum}}_{i=1}^{M}|| FC(h_{v_i}) - \hat{E}_v(v_i)||^2$   
    \end{center}
    \item \textbf{Masked Region Classification (MRC)} aims to predict the semantic class of the masked image region. The transformer prediction is compared with the output of an object detector, by considering the highest confidence object label which serves as the actual target. The cross-entropy (CE) loss between the object detector label $ c(v_i)$ and the transformer prediction $FC(h_{v_i})$ for $m$ regions needs to be optimized:
    \begin{center}
           $\mathcal{L}_{MRC}$ = -$\mathbb{E}_{(\mathcal{W}, \mathcal{V}) \in D}\mathlarger{\mathlarger{\sum}}_{i=1}^{M}
            CE (softmax(FC(h_{v_i}), c(v_i))$  
    \end{center}
    An extension of MRC considers the overall distribution of the object detector label predictions instead of exclusively focusing on the top class. In that case, the objective function attempts to minimize the distance between the object detector distribution and the transformer's distribution of predictions, which is actually equivalent to minimizing the KL divergence between those two distributions. The objective function for \textbf{Masked Region Classification with KL-Divergence (MRC-KL)} for a distribution of object detector labels $\Tilde{c}(v_i))$ instead of the top object detector label $c(v_i)$ can be written as:
    \begin{center}
           $\mathcal{L}_{MRC\textit{-KL}}$ = -$\mathbb{E}_{(\mathcal{W}, \mathcal{V}) \in D}\mathlarger{\mathlarger{\sum}}_{i=1}^{M}
            D_{KL} ((FC(h_{v_i}), \Tilde{c}(v_i))$   
    \end{center} 
\end{itemize}

\textbf{\textit{Image-Text Matching (ITM)}} enables learning visiolinguistic matches in a global level. It can be viewed as a multimodal extension of next sentence prediction (NSP), where the model needs to recognize whether a given pair of text and image is in fact matched or not, as both positive and negative pairs are sampled. The alignment probability between text and image is provided by a score function $s_{\theta}$, and the binary cross-entropy (CE) needs to be optimized: 

\begin{equation*}
    \mathcal{L}_{VLM} = -\mathbb{E}_{(\mathcal{W}, \mathcal{V}) \in D}[y\log s_{\theta}(\mathcal{W}, \mathcal{V}) +  
    (1-y)\log(1 - s_{\theta}(\mathcal{W}, \mathcal{V}))] 
\end{equation*}
\hfill \break
\textbf{\textit{Word-Region Alignment (WRA)}} is a more fine grained version of ITM, where words have to be grounded to image regions.  \\

\textit{Contrastive objectives} act upon data pairs projected to the same semantic space, so that the model learns a representation based on their similarity. \\
\textbf{\textit{Cross-Modal Contrastive Learning (CMCL)}} learns to place matching image-text pairs close together, while pushing apart any mismatched ones. The contrastive loss for the i-th and j-th pairs sampled from $\mathcal{D}$ is mathematically expressed as:

    \begin{center}
    $\mathcal{L}_{CMCL}$ = $- \mathlarger{\frac{1}{n} \mathlarger{\sum}}_{i=1}^{M}\log \frac{\mathlarger{\sfrac{exp(v_i^\top w_i}{\sigma})}}{\mathlarger{\sum}_{j=1}^{M}\mathlarger{\sfrac{exp(v_i^\top w_i}{\sigma})}}$ \\
    \end{center}
 \hfill \break   
Where $v_i$ refers to the image of the i-th pair and $w_j$ refers to the text of the j-th pair, $s_{\theta}(v_i, w_j) = v_i^\top w_j$ is a scoring function which is maximized when matching pairs occur, and $\sigma$ serves as a learnable temperature parameter.
\subsubsection{Task-specific fine-tuning}
\label{sec:finetuning}
Current VL tasks are usually created by extending existing tasks in NLP or vision domains. They may be either  \textbf{\textit{discriminative}} or \textbf{\textit{generative}}, and usually there are variants addressing both problems \cite{survey3}.

\paragraph{Visual Question Answering (VQA)} Given an image $I \in \mathcal{V}$ and a natural language question $q \in \mathcal{W}$, a VQA model is tasked to predict the correct answer $a \in \mathcal{W}$. VQA is an extension of NLP question answering (QA) to include the visual modality. It can be viewed as a \textit{classification} task, where the predicted answer can be selected among a set of candidate answers. Alternatively, free-form answers can be generated, forming a \textit{generative} VQA task.  Widely used datasets for VQA, containing complex scenes of objects and relationships, are the original VQA \cite{vqa}, VQAv2 \cite{vqav2}, GQA \cite{gqa}, Visual7W \cite{Visual7W}, Visual Genome QA \cite{visualgenome} and COCO QA \cite{coco_qa}. Additionally, datasets providing explanations for answer selections, such as VQA-E \cite{e-vqa} have recently emerged. 

\paragraph{Visual Entailment (VE)} VE extends the default task of textual entailment by answering whether an image $I \in \mathcal{V}$ acting as the premise semantically entails the given textual hypothesis $h \in \mathcal{W}$. The hypothesis \textit{h} can either \textit{entail}, \textit{contradict} or remain \textit{neutral} with respect to the premise, providing an answer \textit{a} to the visual premise \textit{I}. SNLI-VE \cite{snli-ve} is a widely used VE dataset. Moreover, e-SNLI-VE \cite{e-SNLI-VE} corrects label errors present in SNLI-VE due to its automatic assembling while providing human-written explanations in natural language for the corrected SNLI-VE corpus.

\paragraph{Visual Referring Expressions (VRE)/Visual Grounding (VG)} 
VG typically focuses on associating textual descriptions $s \in \mathcal{W}$ with specific objects or regions in an image $I \in \mathcal{V}$, aiming to identify and localize objects or scenes mentioned in the text. On the other hand, VRE (as an extension of purely linguistic referring expressions \cite{survey3}) involves generating or retrieving textual descriptions that refer to specific visual elements or regions within an image, aiming to describe and identify those elements in natural language. In essence, VG deals with the interpretation of existing textual descriptions in relation to visual content, while VRE involves generating or retrieving textual descriptions based on visual cues.\footnote{Notably, VG is exclusively discriminative, since image regions are retrieved but not generated, while VRE can be either discriminative (if referring expressions are retrieved) or generative (if referring expressions are generated). VG and VRE are often not discriminated in literature \cite{survey3}. We place the two tasks together, especially since corresponding datasets contain the necessary annotations to either ground visual aspects to text or vice versa.}
Some datasets suitable for VG and VRE are Flickr30k Entities \cite{plummer2016flickr30k}, CLEVR-Ref$+$ \cite{clevr-ref-plus}, RefCOCO, RefCOCO+, RefCOCO-g \cite{coco-ref} and GuessWhat \cite{guesswhat}.

\paragraph{Visual Dialog (VD)} VD is the analogue of chatbots, aiming to maintain a meaningful conversation by responding to consecutive textual inputs $q \in \mathcal{W}$. VD is tasked to create such a dialog upon a given image $I \in \mathcal{V}$. VisDial \cite{visdial} is a dataset for VD that was proposed together with the introduction of the task. Other datasets for multimodal dialog are GuessWhat \cite{guesswhat} and CLEVR-Dialog \cite{CLEVRDialog}.

\paragraph{Image retrieval from text or Text-Image Retrieval (TIR)} TIR attempts to return the most suitable image $I \in \mathcal{V}$ within a database according to a natural language description $c \in \mathcal{W}$. There is also the inverse task of \textbf{text retrieval from image or Image-Text Retrieval (ITR)} that searches the optimal $c \in \mathcal{W}$ given an image $I \in \mathcal{V}$. Cross-modal retrieval, referring to retrieving any modality from the other one is an extension of the NLP task of document retrieval. Common datasets used in TIR/ITR are COCO \cite{coco} and Flickr8k/30k \cite{flickr} which contain images paired with 5 captions each. Variation of cross-modal retrieval tasks involve ambiguous textual statements. For example, \textbf{Visual Word Sense Disambiguation (VWSD)} \cite{raganato-etal-2023-semeval} aims to retrieve an appropriate image $I$ among a set of candidates $V$ based on an ambiguous textual phrase $c \in \mathcal{W}$.

\paragraph{Image Captioning (IC)} IC is a generative task, extending natural language generation (NLG) to describe images: given an image $I \in \mathcal{V}$, provide a sentence $c \in \mathcal{W}$ that describes it. IC can be viewed as a generative counterpart of text retrieval from image (ITR). Dense Captioning is a fine-grained analogue of IC that requires generation of descriptions for image regions instead of providing a global visual caption. Conceptual Captions (CC) \cite{cc},  SBU \cite{sbu}, COCO \cite{coco} and Flickr8k/30k \cite{flickr} are widely used datasets for image captioning. 

\paragraph{Visual Storytelling (VIST)} VIST is the extension of image captioning to a sequence of \textit{N} related images \textit{$I_1, I_2, ..., I_N \in \mathcal{V}$}. Generated captions \textit{$c_1, c_2, ..., c_N \in \mathcal{W}$} should be consistent with each other throughout the sequence, forming a textual 'story'. Datasets related to VIST are Visual Storytelling Dataset (VIST) \cite{vist} which models social language regarding visual concepts,  and New York City Storytelling (NYC-Storytelling) \cite{ny-story}, which contains narratives from blogs.

\paragraph{Multimodal Machine Translation (MMT)} MMT is assisted from the visual modality to translate between two languages, as an extension of the machine translation task. Multi30K-MMT \cite{Multi30K} contains multilingual descriptions of images in English, German, French, and Czech languages.

\paragraph{Visual Reasoning (VR)} VR extends visual perception tasks, such as object detection and classification, semantic segmentation and others. VR needs to predict meaningful relationships between image entities, which is similar to creating a scene graph. Compositional reasoning refers to the task where attributes need to be combined so that the identity of the whole can be inferred. Popular datasets for VR are Compositional Language and Elementary Visual Reasoning (CLEVR) \cite{clevr}, Relational and Analogical Visual rEasoNing (RAVEN) \cite{raven}, Natural Language Visual Reasoning (NLVR) \cite{nlvr} and Natural Language Visual Reasoning for Real (NLVR2) \cite{nlvr2}. In order to test the ability of VR models in novel attribute combinations, CLEVR-CoGenT dataset was proposed as an extension of CLEVR. Real world compositional questions of GQA \cite{gqa} can also be utilized for VR models. Finally, AQUA \cite{aqua} is a visual reasoning dataset dedicated to the artistic domain.

\paragraph{Visual Commonsense Reasoning (VCR)} VCR attempts to understand an image $I \in \mathcal{V}$ by incorporating commonsense knowledge relationships to explain the answer $a \in \mathcal{W}$ derived for each question $q \in \mathcal{W}$. It can be also viewed as an extension of the VQA task, where instead of merely providing an answer \textit{a} to a given visual question \textit{q}, a rationale $r \in \mathcal{W}$ justifying the choice of the answer needs to be returned as well. Usually, $r$ contains commonsense information which is not explicitly shown on the image, but can be inferred based on prior human knowledge. The answer \textit{a} and rationale \textit{r} can be also generated, thus yielding the Visual Commonsense Generation (VCG) task. Widely used datasets are VCR \cite{vcr} and Visual COMmonsense rEasoning in Time (Visual COMET) \cite{visualcomet}.

\paragraph{Vision-and-Language Navigation (VLN)} VLN can be the equivalent of either visual navigation or linguistic navigation in the multimodal domain. Some datasets for VLN are Cooperative Vision-and-Dialog Navigation (CVDN) \cite{cvdn}, Action Learning From Realistic Environments and Directives  (ALFRED) \cite{alfred} and others.

\subsection{Image generation}
There is a lot of architectural variation when the modality to be generated is the visual one, greatly diverging from architectures employed for \textbf{\textit{discriminative}} visual tasks. Image generation can be performed by leveraging Generative Adversarial Networks (GANs), Vision Transformers, Diffusion models or a combination of them.

\subsubsection{Conditional image generation tasks}
\paragraph{Conditional Image Generation (cIG)} CIG addresses the synthesis of an image \textit{I} guided by textual information $s \in \mathcal{W}$, extending uncoditional image generation. Traditionally, text to image generation is performed by conditional GANs (cGANs) \cite{cgan}, which attempt to generate realistic images semantically corresponding to a text description. Image generation from text can be considered as the generative counterpart of \textbf{image retrieval from text (TIR)}, and can be also regarded as the inverse task of \textbf{image captioning (IC)}. Text to image GANs have been benchmarked on a plethora of datasets, either containing simpler images with simple conditioning such as ImageNet \cite{imagenet}, Oxford Flowers \cite{oxfordflowers}, FFHQ \cite{ffhq} and CIFAR \cite{cifar}, longer conditioning such as captions in natural language, as in CUB \cite{cub, t2i}, or even more complex scenes accompanied with captions as conditioning, such as COCO \cite{coco}.

\paragraph{Story Visualization (SV)} SV refers to the synthesis of a visual sequence \textit{$I_1, I_2, ..., I_N$} based on an input story \textit{$c_1, c_2, ..., c_N \in \mathcal{W}$}, the inverse task of \textbf{visual storytelling (VIST)}. Once again, GAN architectures are leveraged to produce the sequence of images, which should not only be realistic, but also maintain the serial progression from frame to frame while remaining relevant to their textual description. Datasets accompanying SV research are CLEVR-SV \cite{storygan},  Pororo-SV \cite{pororo-sv}, Flinstones-SV \cite{flinstones-sv} and DiDeMo-SV \cite{didemo-sv}. 

\subsubsection{Generative VL architectures}
\paragraph{Text-to-image synthesis}
Adversarial text to image synthesis has demonstrated a long line of impressive results by converting textual inputs such as captions, interactive dialogs, sequential story-like captions or structured formats of textual inputs (such as scene graphs and layouts) to plausible images. Some of the first ventures \cite{t2i1, t2i2} achieve synthesizing images, even though their quality is fairly low. To resolve this limitation, subsequent implementations synthesize images in stages, increasing image resolution step by step. StackGAN \cite{stackgan} and StackGAN$++$ \cite{zhang2018stackgan} exploit stacked generators and discriminators, dedicated to coarser or finer resolutions. AttnGAN focuses on individual words rather than whole sentences to synthesize finer details of the image \cite{xu2018attngan}. This idea is extended by SEGAN \cite{segan}, which attends on relevant keywords from the sentence rather than all existing words. The most significant parts of the sentences with respect to the image contribute in synthesizing images with reduced fuzziness in details, as reported in DM-GAN \cite{zhu2019dmgan}.

\paragraph{Sequential text-to-image synthesis}
StoryGAN \cite{storygan} is the model that introduced the Story Visualization (SV) task, which concerns synthesizing a sequence of related images, maintaining consistency across story frames. The main StoryGAN architecture consists of a generator and two discriminators (image and story discriminators) guided by an RNN structure responsible of encoding the textual story. Improvements on the basic model are performed in \cite{pororogan, li2020storygan}. More recent models substitute the RNN encoding scheme with transformers \cite{Maharana2021ImprovingGA, Maharana2021IntegratingVL, character-centric, impartial, cmota, papadimitriou2024masked}, following the same trend as in other multimodal tasks. 

\paragraph{Text-to-image generative VL Transformers}
Surprisingly, VL transformers in their original form are not able of generating realistic images, despite their impressive capabilities in other visual tasks. This issue is attributed to the fact that regression based training objectives, such as the one used in LXMERT \cite{tan2019lxmert}, are not able to handle feature generation in high dimensional spaces.
Getting VL transformers one step further, X-LXMERT \cite{xlxmert} as an extension of LXMERT, generates caption-conditioned high fidelity images consistent to their descriptions.  X-UNITER follows the same extension logic as X-LXMERT, based on the UNITER architecture \cite{chen2020uniter}. The image generation capabilities of X-UNITER are comparable to X-LXMERT, showcasing the general applicability of this approach. 

DALL-E \cite{dalle} uses a 12 billion parameter version of GPT-3 to generate fine-grained and highly diverse images based on corresponding textual descriptions. It demonstrates a large range of conditional synthesis capabilities, such as controlling certain visual attributes, as well as object positioning in an accurate way, capturing and visualizing 3D scenes, performing several natural effects like reflections, inferring missing details from descriptions, combining unrelated concepts in one image, performing zero-shot reasoning in the visual domain, incorporating external spatial and temporal knowledge and others.  Once again, a connection between model scale and advanced performance is observed in terms of zero-shot generation and regarding the range of generalization capabilities. 

\paragraph{Text-to-image diffusion models} Diffusion models are setting new baselines for state-of-the-art conditional image generation \cite{diffusion}. They follow a process of gradually adding Gaussian noise on an image and then learn to reconstruct it. During the last year, the field of diffusion-based image synthesis has received a variety of interesting works. 
Stable Diffusion \cite{diff0} enables previously computationally demanding image synthesis even within limited resources scenarios by applying the diffusion training process in the latent space of autoencoders rather than pixel-level operations in the image space. 
DALL-E2 \cite{dalle2} extends the high-quality result of its predecessor \cite{dalle} by using learned text-conditioned image embeddings obtained from CLIP \cite{clip} as conditionings to a diffusion model acting as the decoder. Synthesized images are photorealistic and faithful to their conditioning, while zero-shot language-guided manipulation on a source image is also possible.
The concurrent work of Imagen \cite{diff1} exploits large pre-trained language models, such as T5 \cite{t5} for language encoding and proceeds with image synthesis based on the diffusion process.
DreamBooth \cite{diff2} steps upon Imagen to contextualize image synthesis, given variable context described in text. Therefore, different variations of visual subjects can be obtained, maintaining high synthesis quality. Currently, the diffusion-based literature in text-to-image remains abundant and rapidly evolving \cite{tang2023iterinv, yuan2023customnet, circumvent, kim2023wouaf, segalis2023picture, vuong2023languagedriven, patel2024conceptbed}.

\paragraph{Combined GAN/Transformer text-to-image models}
Hybrid architectures utilize existing GAN generators together with powerful VL transformers such as CLIP \cite{clip}. Given an input text prompt, CLIP is responsible of guiding image synthesis in the latent space, based on the text-image similarities it has learned. FuseDream \cite{fusedream} follows this paradigm by optimizing the latent space of pre-trained GANs for efficient navigation. Similar hybrid text to image implementations are Big Sleep\footnote{\href{https://colab.research.google.com/github/levindabhi/CLIP-Notebooks/blob/main/The_Big_Sleep_BigGANxCLIP.ipynb}{BigSleep notebook}} and VQGAN$+$CLIP \footnote{\href{https://colab.research.google.com/github/justinjohn0306/VQGAN-CLIP/blob/main/VQGAN\%2BCLIP(Updated).ipynb}{VQGAN+CLIP notebook}}. 

\subsection{Evaluation metrics for VL models}
\subsubsection{Classification metrics}
\textbf{\textit{Discriminative}} VL tasks that provide an answer selected among pre-defined candidates are evaluated via \textit{classification} metrics. Such tasks include visual question answering (VQA), visual grounding (VG), visual reasoning (VR), visual entailment (VE).
\textbf{Accuracy@k} measures the proportion of correct answers over all answers, where an answer is considered to be correct if it belongs within the top-k ones. It can serve as a generic measure of quality for \textbf{\textit{discriminative}} VL tasks, 
irrespectively of the modality that is predicted as the answer: for example, in the case of VQA, accuracy refers to the linguistic modality, as a textual answer needs to be selected. 
When bounding boxes have to be predicted, as in the case of VG, \textbf{Intersection over Union (IoU)} provides a measure of success regarding the overlap between ground truth and predicted bounding boxes. Higher values are better, indicating a larger percentage of overlap.

\subsubsection{Ranking metrics}
\textit{Ranking metrics} \cite{ir} provide further insights regarding the success of retrieving the right answer by providing information related to the position the ground truth answer was ranked. Tasks commonly using ranking metrics are image-text \& text-image retrieval (ITR/TIR), visual storytelling (VIST), visual dialog (VD), machine translation (MT).
\textbf{Recall@k (R@k)} measures the proportion of total ground truth instances that were found in the top-k rank, without taking into account their ordering. Higher Recall@k scores are better.
\textbf{ Precision@k (P@k)} is the percentage of ground truth answers in the top-k over all retrieved top-k items, without taking into account their ordering. Higher Precision@k scores are better.
When ordering is to be considered, \textbf{Mean Reciprocal Rank@k (MRR@k)} acts as a useful performance measure. As reciprocal rank of an answer is considered the inverse of the rank position of the -first- right answer. By averaging over all instances, MRR is derived, with higher values indicating better performance. For $N$ instances, if $rank_i$ denotes the position of instance $i$, MRR can be written as:
\begin{center}
        $MRR$ = $\mathlarger{\frac{1}{N}\mathlarger{\sum}}_{i=1}^{N} \frac{\mathlarger{1}}{\mathlarger{rank_i}}$
\end{center}
\hfill\break
Another order-sensitive metric is \textbf{Discounted Cumulative Gain (DCG)}, which measures the gain an answer offers based on its ranking by taking into account a graded relevance scale. The graded relevance value is supposed to reduce logarithmically with respect to the position, thus penalizing highly relevant answers that appear lower in the rank. For $rel_i$ the graded relevance at position $i$ and $p$ a particular rank position, $DCG$ score for $p$ is defined as:

\begin{center}
        $DCG_p$ = $\mathlarger{\mathlarger{\sum}}_{i=1}^{p} \frac{\mathlarger{rel_i}}{\mathlarger{\log_2(i+1)}}$
\end{center}
\hfill\break
\textbf{Normalized Discounted Cumulative Gain (NDCG)} is a normalized rank quality score that represents the ratio between the $DCG$ score of the rank returned by an algorithm divided by the ideal order $DCG$ ($iDCG$).

\textbf{Median Rank (MedRank)} is the median position of ranked ground truth answers when all answers have been considered. In a similar fashion,
\textbf{Mean Rank} denotes the average position of ranked ground truth answers when all answers have been considered. Lower median and mean rank values are better.

\subsubsection{Similarity metrics}
\paragraph{Wu-Palmer similarity (WUPS)} \cite{wups} reports the degree of similarity between two words based on their least common subsumer on WordNet \cite{wordnet} taxonomy. Various thresholds can define the agreement between two candidates, with the most typical being 0.0 and 0.9.

\subsubsection{Language generation metrics}
Language generation tasks such as Image Captioning (IC), VIsual Storytelling (VIST)  and generative versions of Visual Question Answering (VQA), Visual Reeferring Expressions (VRE) and Visual Commonsense Reasoning (VCR - VCG) evaluate produced outputs using the following \textit{language generation metrics}.

One of the oldest metrics is \textbf{Bilingual Evaluation Understudy (BLEU)} \cite{bleu}, originally developed to evaluate machine translation. It compares how much a machine generated linguistic output matches human-written text in a precision-oriented way. Therefore, it is supposed to demonstrate high agreement with human perception regarding the quality of generated text, with higher BLEU score indicating higher perceived quality and maximum BLEU score being equal to 1 (or 100\%). BLEU  not only measures individual word matches, but also n-gram (grouped words) matches, providing BLEU-N scores for different N (most usually N = 1 for unigrams, 2 for bigrams, 3 for trigrams, 4 for quadrigrams). Nevertheless, BLEU cannot assess linguistic diversity of generated text and sometimes it is unable to appropriately perform in practical settings. BLEU is rather reliable for long sentences, but not very helpful in short/monolectic ones.

\textbf{Recall Oriented Understudy for Gisting Evaluation (ROUGE)} \cite{rouge} is designed to evaluate the quality of summarization compared to a human-made summary. Similarly to BLEU, n-grams of varying N are utilized to calculate ROUGE-N scores. ROUGE-N offers not only recall, but also precision and F1 evaluation. Specifically, \textbf{ROUGE-N Precision} measures how many overlapping n-grams were found over the total generated n-grams: \\

    \begin{center}
        $\textit{ROUGE-N}_{precision}
        =\mathlarger{\frac{\textit{common\ n-grams\ in\ generated\ and\ reference}}{\textit{n-grams\ in\ generated}}}$ 
    \end{center}
\hfill\break    
 \textbf{ROUGE-N Recall} assesses how many overlapping n-grams were found compared to the human-made reference: \\
 
    \begin{center}
        $\textit{ROUGE-N}_{recall}=
        \mathlarger{\frac{\textit{common\ n-grams\ in\ generated\ and\ reference}}{\textit{\ n-grams\ in\ reference}}}$ \\
    \end{center}
\hfill\break    
Finally, \textbf{ROUGE-N F1 score} takes into account both ROUGE-N precision and recall: \\

    \begin{center}
        $\textit{ROUGE-N}_{F1}=
        2 * \mathlarger{\frac{(\textit{ROUGE-N}_{precision}) * (\textit{ROUGE-N}_{recall})}{(\textit{ROUGE-N}_{precision}) + (\textit{ROUGE-N}_{recall})}}$ \\
    \end{center}
\hfill\break    
Additionally, there is \textbf{ROUGE-L} score, which measures the longest common subsequence (LCS) between generated and ground truth text, with a longer LCS implying higher similarity. ROUGE-L can operate in sentence-level or summary-level and demonstrates precision, recall and F1-score variants. Specifically, \textbf{ROUGE-L Precision} measures the LCS length compared to generated n-grams count: \\

    \begin{center}
        $\textit{ROUGE-L}_{precision}=\mathlarger{\frac{\textit{LCS (common\ n-grams)}}{\textit{count\ of\ n-grams\ in\ generated}}}$ 
    \end{center}
\hfill\break    
Moreover, \textbf{ROUGE-L Recall} measures the LCS length compared to ground truth n-grams count: \\

    \begin{center}
        $\textit{ROUGE-L}_{recall}=\mathlarger{\frac{\textit{LCS (common\ n-grams)}}{\textit{count\ of\ n-grams\ in\ reference}}}$ 
    \end{center}
\hfill\break    
\textbf{ROUGE-L F1 score} takes into account both ROUGE-L precision and recall: \\

    \begin{center}
        $\textit{ROUGE-L}_{F1}=2 * \mathlarger{\frac{(\textit{ROUGE-L}_{precision}) * (\textit{ROUGE-L}_{recall})}{(\textit{ROUGE-L}_{precision}) + (\textit{ROUGE-L}_{recall})}}$ 
    \end{center}
\hfill\break

There are also some other rarely used variants of ROUGE score. \textbf{ROUGE-W} searches for the Weighted Longest Common Sub-sequence. \textbf{ROUGE-S} (Skip-Bigram Co-Occurrences Statistics) measures the overlap of non-consecutive n-grams between generated and reference sequences. Similar to BLEU, ROUGE scores are more effective in long sentences rather than short ones.

\textbf{Metric for Evaluation of Translation with Explicit Ordering (METEOR)} \cite{meteor} tackles the need for exact word matching as in BLEU score, and instead enforces semantic matching, taking into account possible synonyms and paraphrases of words in the reference text. Semantics' matches are made possible due to the usage of WordNet \cite{wordnet}. Unigram alignment between ground truth and generated sequences contribute to the METEOR score. Multiple possible alignments between the two sequences are resolved by selecting the alignment with the fewest crosses, when ground truth and generated unigrams are matched. METEOR presents high agreement with human perception regarding the similarity of generated text sequences.  

\textbf{Consensus-based Image Description Evaluation (CIDEr)} \cite{cider} is a metric inspired from human agreement when ground truth and generated sentences are compared. Such similarity can be expressed via TF-IDF score of n-grams across reference sentences, instructing frequently occurring n-grams in the whole dataset to have lower weight in the final score, as it is more possible to be less informative (IDF term), while at the same time increasing the weight of frequent n-grams within a reference sentence (TF term). Similar to METEOR, CIDEr exploits semantic matching by comparing stemmed versions of words in sentences. A cosine similarity score between generated and ground truth vectors computed from TF-IDF n-gram weights provides the final CIDEr score.

\textbf{Semantic Propositional Image Captioning Evaluation (SPICE)} \cite{spice} is an automated evaluation metric for language generation operating over scene graphs, which are synthesized from ground truth captions and generated captions via dependency parsing. WordNet is utilized for disambiguation and synonym detection. Even though it resolves shortcomings of previous methods related to n-gram overlaps, the scene graph construction stage is possible to induce errors early in the evaluation process.

\textbf{BLEURT} \cite{bleurt} is a BERT-based learned evaluation metric that addresses the shortcomings of traditional BLEU and ROUGE metrics in order to better correlate the resulting scores with human perception. It combines the merits of learning pure linguistic associations, as well as human assessments over language metrics. Therefore, pre-training on synthetic sentence pairs, which are designed to capture semantic, syntactic and lexical information assists in identifying possible dissimilarities between real and generated text. Afterwards, fine-tuning adapts automatically captured disagreements to actual human ratings.

\subsubsection{Image generation metrics}
\label{sec:imgen_metrics}
\textbf{Inception Score (IS)}\cite{is} evaluates the quality and diversity of generated images by utilizing a pre-trained image classifier, such as Inception V3 \cite{inceptionv3}. The classifier is tasked to provide a probability of whether an image is generated or real. IS can have a lowest value of 1.0 and a highest value equal to the number of classes the pre-trained classifier has seen, which in case of Inception V3 is 1000 (ImageNet \cite{imagenet} classes). Higher IS implies better quality and more diversity, therefore higher IS is better.

\textbf{Fréchet Inception Distance (FID)}\cite{fid} compares the distribution of generated images with the distribution of the real ones by taking into account the mean and the standard deviation of the distributions. A pre-trained Inception-V3 \cite{inceptionv3} model is utilized to extract summary statistics for real and generated images. Lower FID scores are better, indicating higher similarity between generated and real images.

\textbf{Learned Perceptual Image Patch Similarity LPIPS)} \cite{lpips} is a metric designed to reflect the perceptual similarity between real and generated images using deep features of image classifiers. Lower LPIPS values indicate more similar images.

\textbf{R-precision} is another widely used metric to evaluate synthesis quality. Despite primarily being a retrieval metric, it can serve conditional generative models: generated images are used as queries to ground truth descriptions, with higher scores indicating better quality of generated images.

\section{Graphs}
\label{sec:graphs}
A graph structure $\mathcal{G} = \mathcal{V}, \mathcal{E}$ consists of a set of nodes $\mathcal{V}$ interconnected with weighted or unweighted edges from a set $\mathcal{E}$. Edges also can be either directed or undirected depending on the constrains of relationships they express. Considering two nodes $v_i \in \mathcal{V}, v_j \in \mathcal{V}$, their in between edge is denoted as $(v_i, v_j) \in \mathcal{E}$. Nodes and edges can additionally contain features. There are some distinct subcategories of graphs mentioned below, often present in real world data representation scenarios \cite{graph-book}.

Different types of relationships raise the need for distinct edge representations, which is satisfied via \textbf{Multi-relational graphs}. The edge notation needs to be altered to contain the edge type $\tau$, so that the multi-relational edge notation being $(v_i, \tau, v_j) \in \mathcal{E}$.
\textbf{Heterogeneous graphs} extend multi-relational graphs by introducing varying node types, so that nodes can contain labels forming non-overlapping node sets. Therefore, the node set $\mathcal{V}$ can be expressed as a union of node labels $\mathcal{V} =\mathcal{V}_1 \cup \mathcal{V}_2 \cup ... \cup \mathcal{V}_n$ for $n$ distinct and disjoint node categories. Those categories most often impose constrains on the edge types as well in order to remain meaningful. 
\textbf{Multipartite graphs} are heterogeneous graphs which exclusively contain edges that connect nodes of different types.

\subsection{Knowledge Graphs}
A Knowledge Graph (KG) $\mathcal{G}$ is a structured representation of facts $\mathcal{F}$, which consist of entities $\mathcal{E}$, relationships $\mathcal{R}$ and semantic descriptions. Entities can describe either abstract concepts or actual objects, relationships form the meaningful connections between entities, and semantic descriptions incorporate types and properties of those objects and relationships. KGs are directed and heterogeneous structures that describe human knowledge in the form of triplets \textit{head - relationship - tail}, often denoted as \textit{(h, r, t)} $\in \mathcal{F}$, or equivalently \textit{subject - predicate - object}, denoted as \textit{(s, p, o)} $\in \mathcal{F}$. \cite{kg1}

While existing edges express known facts, there are two scenarios for missing edges \cite{kg1}: Open World Assumption (OWA) assumes that unobserved facts are either missing or false;  Closed World Assumption (CWA) assumes that all unobserved facts are false.

\subsection{Graph representation}
Graph representation has been a field of increasing interest, as it affects numerous applications in artificial intelligence. Similar to language representations, KG embeddings are low-dimensional mappings of the graph entities and relationships. These vector representations capture the semantic information contained in KGs, which can consequently be used for a variety of downstream tasks \cite{kg1, ge1}.

Node embeddings constitute a family of graph embedding algorithms aiming to represent the nodes' position and context in a vector space. Popular methods regarding node embeddings are based on random walks, presenting widely used implementations such as node2vec \cite{node2vec}, DeepWalk \cite{deepwalk}, Large-scale Information Network Embedding (LINE) \cite{line}, Graph2vec \cite{graph2vec} and others.
However, standalone node embeddings are inadequate for multi-relational graph representations, which require an overall representation of nodes, edges and attributes. To this end, shallow translational models exploit geometric capabilities of the vector space to achieve multi-relational graph representations \cite{graph-book}. 

\textbf{Graph Neural Networks (GNNs)} refer to a broad framework that achieves graph structure and feature representations using deep learning. The basic idea behind GNNs is neural message passing, i.e. the exchange and aggregation of neighboring node information through their connections, resulting in updating the node embedding itself. The process repeats by expanding to more distant neighborhoods (hops) in every step, integrating more information regarding graph structure and neighboring features \cite{graph-book}. 

More specifically, each node $v \in \mathcal{V}$ of a graph $\mathcal{G}$ is initialized with a random embedding $h_0$ and the following hidden embedding states $h_v$ are computed via a local transition function \textit{f} such that:
\begin{equation}
    h_v = f(x_v, x_{co[v]}, h_{N_v}, x_{N_v})
\end{equation}
where $x_v$ are the features of the node $v$, $x_{co[v]}$ are the edge features connected to $v$, $h_{N_v}$ are the embedding states of the neighbors of $v$ and $x_{N_v}$ are their features.
The output is given from a local output function \textit{g}:
\begin{equation}
    o_v = g(h_v, x_v)
\end{equation}

\textbf{Graph Convolutional Networks (GCNs)} are a variant of CNNs operating on graphs. However, different than images, graphs contain unordered nodes with varying numbers of connections. Given a graph $\mathcal{G} = (\mathcal{V}, \mathcal{E}, \mathcal{X})$, the GCN receives as an input the feature matrix $\mathcal{X}$ containing node features, and an adjacency matrix $\mathcal{A}$ representing the graph structure. The convolution operation is generalized so that the representation of a current node is obtained by aggregating its own features, as well as the features of its neighbors. A non-linear transformation is then applied on the aggregated features. Many approaches stack multiple convolutional layers so that feature information from further neighborhoods can be received \cite{gcn, gnn3}.

GCNs can effectively be applied on relational data and serve a variety of relevant downstream tasks. Relational Graph Convolutional Networks (R-GCNs) comprise a GCN variant for relation-specific encodings for KGs, i.e. encodings depending on edge directivity and type. Therefore, different weights will be assigned to different relationships \cite{rgcn}.

\textbf{Graph Attention Networks (GATs)} are based on attention mechanisms, which have the ability of handling variable sized inputs. Therefore, attention mechanisms could be applied successfully for the representation of graphs containing nodes with different edge degrees. Indeed, self-attention allows each node to attend to its neighbors, thus computing the hidden encoding. This process can be parallelized for neighboring pairs, independently of graph structure \cite{gat}.

Despite GNNs being able to learn representation on fixed and homogeneous graphs, they are not so powerful on graphs containing heterogeneous information on nodes and edges. In such cases, embedding extraction mechanisms should be adjusted appropriately to encode heterogeneous information. Various types of relationships create meta-paths between nodes, containing diverse and rich semantic information. GATs are extended by applying a hierarchical attention mechanism with node-level and semantic-level attention, assigning different importances to nodes and meta-paths. Neighboring features are aggregated in a hierarchical fashion forming the final node embeddings \cite{hgat}.

\textbf{Graph Transformer Networks (GTNs)} can also handle heterogeneity and generate reasonable meta-paths connecting nodes, as well as effective node representation. Meta-paths may be of any length up to the number of Graph Transformer layers
and contain arbitrary edge types. Multiple generated meta-paths can simultaneously be considered, defining multiple learned graphs. Due to continuously generating new graph structures from adjacency matrices contained in data, more powerful node representations are learned through convolutions, as there are higher chances of finding more useful meta-paths. A GCN is applied on each meta-path, yielding an ensemble of GCNs when considering all the generated meta-paths \cite{gtn}.

The Heterogeneous Graph Transformer (HGT) is designed to model web-scale heterogeneous graphs using heterogeneous attention. Node and edge-level dependent parameters are exploited in order to obtain dedicated node and edge representations. HGT extracts all related pairs from a sampled heterogeneous subgraph, so that each pair contains a source node and a target node. The information from source nodes is aggregated to provide a contextualized representation for a target node. This process can be decomposed in three parts, namely Heterogeneous Mutual Attention, Heterogeneous Message Passing and Target-Specific Aggregation. The weight matrices for the heterogeneous mutual attention, message passing, and propagation steps are parameterized using the meta-relations of the graph. Dynamic representation graphs are handled using the relative temporal encoding technique, which captures arbitrary long dynamic structural dependencies \cite{hgt}.

\subsubsection{Graph representations in KVL models}
Apparently, one important aspect towards KVL models is how knowledge is incorporated with vision and language modalities. First attempts to incorporate knowledge stored in KGs were based on exact matching between extracted visual and textual concepts from the given input and existing KG nodes.  
However, exact matching will result in errors even when there is a small discrepancy between KG entities and extracted concepts, limiting the contributions of additional knowledge. In order to better exploit semantically rich structured knowledge sources, more refined strategies, such as the usage of GNN representations are explored in recent literature. The variability of GNN implementations as described previously allows dedicated representations for different graphs, enabling the incorporation of node, edge and feature information. Therefore, retrieved knowledge is more accurate and informative, ultimately capable of boosting VL models.

\section{Knowledge}
\label{sec:knowledge}
In recent years, there has been an ever increasing interest in incorporating external knowledge \textit{K} in VL models. We can identify the merits of such an approach: additional knowledge can offer performance boosting, extendability and potentially explainability of existing VL tasks. Most common senses of additional knowledge offering those benefits are analyzed below. 

\textbf{Hierarchical knowledge} refers to \textit{is-a} relationships forming a tree structure, with the root serving as the most generic concept and parent node of all the rest, while leaves constitute the most specific concepts. For example, \textit{cat is-a mammal} is an instance that represents such hierarchical relationships.

\textbf{Lexical knowledge} serves as a structured dictionary, offering linguistic rules, while being able to resolve issues such as word sense disambiguation. Lexical knowledge can be combined with hierarchical knowledge, providing hypernym/hyponym relationships.

\textbf{Named entities} cover a variety of proper names as instances of entities, and include names of people, locations, companies, organizations etc \cite{named_entities}. For example, the sentence \textit{Joe Biden is the president of the United States} contains the named entities \textit{Joe Biden} and \textit{United States}.

\textbf{Factual knowledge} includes encyclopedic information of the world, such as the historical fact \textit{WW2 lasted from 1939 to 1945}. Such knowledge can also refer to more specific scientific facts, including knowledge in medical, biological, chemistry domains and many more. Facts can also be combined with named entities, forming statements such as \textit{Zebras live in Africa} (\textit{Africa} is a named entity).

\textbf{Commonsense knowledge} is the self-evident perception of the world according to humans; \textit{sugar is sweet} and \textit{if I go out in the
rain I’ll get wet} are obvious commonsense statements. We can identify several discrete senses of commonsense knowledge, affecting aspects of the world a human experiences. Such subcategories refer to \textbf{similarity/dissimilarity} concepts, such as synonyms and antonyms of words. Another commonsense variant includes \textbf{part-whole} (part-of) relationships representing concepts belonging to more generic ones or consisting members of a group, for example \textit{the bark is a part of a tree, the tree is part of the forest}. Part-whole in terms of lexical knowledge is expressed via \textbf{meronyms} (part) and \textbf{holonyms} (whole). \textbf{Utility} relationships describe usage scenarios, such as \textit{a fork is used for eating} or capability (\textit{wheels can rotate}). \textbf{Spatial} information offers knowledge about usual locations of objects in the physical world, for example \textit{boats are situated near water}, or even geographic information, such as \textit{Italy is located at Europe}, which sits on the intersection with \textbf{factual knowledge} and \textbf{named entities}. \textbf{Comparative knowledge} provides rules of comparison between objects, for example \textit{leopards are larger than cats}. Such statements are crucial towards learning logical reasoning scenarios.
\textbf{Numerical knowledge} addresses common enumerations in real life, providing facts such as \textit{humans have two eyes}. \textbf{Intents}, \textbf{desires} and \textbf{plans} constitute another sense of commonsense knowledge, including facts such as \textit{hungry people want to eat} and \textit{a hungry person cooks to eat}. \textbf{Behavioral} knowledge results from logical reasoning over commonsense facts, forming rules e.g. \textit{a child cannot drink 10 liters of water in one day}. \textbf{Creator knowledge} contains statements  such as \textit{a song is created by a musician} or \textit{bread is made from wheat}.

\label{temporal}
\textbf{Event/temporal knowledge} contains chronological information and order of events, blending factual and commonsense knowledge. Events can refer to a large number of chronologically distinct time periods from widely known events such as \textit{world wars, significant political events, sports, social/scientific movements} and many more, to more specialized events known to smaller audiences. Temporal sequences can contain chronologically ordered events: for example, \textit{COVID-19 started in 2019. Vaccines for COVID-19 were developed during 2020} is a \textbf{factual sequence} of events. A \textbf{commonsense sequence} of events could contain information such as \textit{Spring comes after winter}. Sequences may also refer to \textbf{causal} relationships with the cause preceding the event, such as \textit{the boy dropped a glass of water and then the glass broke}, which can also be transformed to \textbf{hypothetical if/then} statements, for example \textit{if a boy drops the glass of water, the glass will break}, or even \textbf{counterfactual} statements expressing what would have happened if an alternative scenario occurred, e.g. \textit{if the boy had not dropped the glass of water, the glass would not have been broken}.

\textbf{Visual knowledge} contains images and possibly additional annotations to connect \textbf{visual perception} with \textbf{commonsense}. Attributes of objects, such as shape, color, texture and others can be connected with their visual counterpart, visualizing commonsense situations such as \textit{tomatoes are red and round}. Visual knowledge is ideal for learning instances of the world involving object relationships and attributes, paving the way for more complex reasoning required in several multimodal tasks. \textbf{Spatial} relationships are naturally combined with images; for example \textit{apples placed inside a bowl, bowl placed on a table}. More types of relationships can be further visualized, including \textbf{actions} between visual entities (\textit{a girl is holding a tennis racket}), object \textbf{details} (\textit{black and white stripped hat}), \textbf{part-whole} has-a relationships (\textit{woman has long blonde hair}) or \textbf{scene text} (\textit{a truck with Coca-Cola logo}). Those rather obvious statements can be extended to commonsense \textbf{assumptions} (\textit{the temperature is low}), when an image of an icy landscape is provided. More complex visual instances can provide information about \textbf{intents} (\textit{a customer enters a restaurant to eat}, \textit{a person holding a suitcase and a passport plans to travel}), \textbf{causes/effects} (\textit{a biker cycling out in the rain will get wet}), \textbf{factual} instantiations (\textit{an ancient Greek temple of the 5th century BC}, \textit{girls with Japanese kimono dresses}), \textbf{similarity reasoning} (\textit{the dog's toy looks similar to a plate}), 
similarity including \textbf{named entities} (\textit{a man looking similar to Brad Pitt}), \textbf{creator knowledge} (\textit{the painting was created by a person holding a paintbrush}), \textbf{capability} (\textit{a cat can jump on the tree branch}).

\subsection{Types of knowledge sources}
We divide \textbf{\textit{external}} knowledge in three categories: \textbf{implicit knowledge}, present in a non-symbolic form, \textbf{explicit knowledge}, typically stored in structured knowledge bases, and \textbf{web-crawled} knowledge, acquired from various online sources, usually in unstructured format. 

Moreover, we can recognize the category of \textbf{\textit{internal knowledge}} or \textbf{\textit{self-knowledge}}, which does not rely on external sources, but rather obtains extra knowledge from the existing data.

\textbf{Implicit knowledge} refers to information stored in a non-symbolic form, such as neural network weights. The indisputable popularity of neural architectures in recent deep learning literature has led to numerous relevant contributions, even if their primary goal deviates from knowledge representation. Unsupervised or self-supervised pre-training of transformer models is able to provide implicit knowledge in several linguistic \cite{implicit} or multimodal tasks. Therefore, incorporating large-scale linguistic and visual data in the pre-training stage can seemingly form unstructured knowledge bases. 

Insightful studies have attempted to discover the optimal pre-training regime and whether it serves as a necessary prior for performance. The right dataset and design choices are crucial for achieving successful representations, which can become a prerequisite for the overall success of the final task, after proper fine-tuning. First, the amount and type of pre-training data need to be examined. Specifically, the relevance of the selected pre-training dataset with respect to the downstream task has been proven more influential rather than dataset size, an observation that remains valid even when generated datasets are used instead of out-of-domain natural ones. Only scaling up in-domain data seems to positively impact the model, encouraging the scalability of implicit knowledge bases. In some cases, especially when selected pre-training datasets are not diverse enough, pre-training transferability towards downstream tasks is low, so that in fact pre-training knowledge is deemed insufficient. In the same fashion, even though the pre-training scenario which demonstrates the lowest losses can be regarded as the best prior, it can be proven suboptimal without proper fine-tuning. Data relevance seems to be more significant than model size, even though deeper single stream models mitigate the semantic gap between images and text in later layers. However, for double-stream models, earlier layers present a more narrow semantic gap, disagreeing with the single-stream observation \cite{survey6, pretraining_right}.

Other investigations question whether all participating modalities contribute equally to the learned representations, proving that during inference, the contribution of language is more prominent than vision in both \textbf{\textit{single-stream}} and \textbf{\textit{double-stream encoder}} architectures. Nevertheless, rich visual knowledge is encoded in pre-training, effectively capturing visual relationships \cite{survey6}.  Such observations are able to offer valuable insights to the complex pre-training procedure and provide enhancements on the knowledge acquired, towards more high-quality, flexible and robust implicit knowledge bases. 

Pre-training holds the advantage that it can exploit unlabelled data for achieving a generic understanding of language \cite{bert, gpt3}, which serves as an initial point for linguistic or VL tasks. In most multimodal cases however, a level of supervision is required, such as the need for paired images and captions. The current abundance of paired image-text samples resolves the labelling issue up to some point, at least for general-purpose domains. The raw nature of linguistic, visual and paired image-text data that are typically used for pre-training alleviates the need for a strict representation, which would limit the flexibility of incorporated knowledge in many aspects, including storage, expressivity and accessibility \cite{implicit}. Furthermore, the automatic incorporation of additional information by repeating pre-training or performing fine-tuning can help extending and refining the required implicit knowledge, contrary to handcrafted knowledge bases which are hard to be extended at scale. Even though such pre-training procedures are very expensive computationally, luckily pre-trained transformer models are offered to the research community in ready-to-use models. Transfer learning can then effectively leverage existing implicit knowledge sources achieving impressing results in various tasks, which accounts for many advancements in VL tasks and transformer-based implementations in general.

Nevertheless, implicit knowledge is not always sufficient to answer questions requiring general, factual and commonsense knowledge, especially when rare information is requested. Additionally, its black-box nature raises concerns about how and what a pre-trained model has learned; for example, biased or erroneous data received during pre-training will be reflected in all later stages, resulting in decreased performance of the downstream VL model. Tracing back the source of such a problem is not possible due to the lack of interpretability tied to implicit knowledge bases.

\textbf{Explicit knowledge} is based on clear, structured facts in the form of a knowledge graph and it is able to explicitly fill the gaps that cannot be covered via transfer learning. Even though most contemporary multimodal approaches, including transformer-based ones, have acquired a certain understanding of language, visual concepts and their in between relationships, they cannot effectively handle concepts and relationships they have never seen during training \cite{commonsense-dimensions} (excluding implementations that present zero-shot capabilities \cite{clip, simvlm}\footnote{In the emergence of Large VL models (analogous to Large Language Models for purely linguistic tasks) more and more implementations may present zero-shot capabilities, even though they are tied with issues such as unfaithful outputs (hallucinations) and increased computational demands. Large VL models are not currently the focus of this paper.}). Consequently, even the best VL models will fail in cases the data distribution is significantly different from the one they have been trained on. The same discrepancy may apply even when an implicit knowledge source is used, if the implicit distribution remains rather distant. For example, a model trained on pairs of generic images and corresponding captions will inevitably present much lower metrics when asked to infer from medical images accompanied by relevant captions with scientific vocabulary. The same limitation is prevalent when there is a shortage of training data \cite{commonsense-dimensions}. Although an intuitive scenario would suggest to repeat the pre-training procedure, so that this extra information will be reflected via updated neural weights, the pre-training cost is in reality computationally unaffordable \cite{pretraining-cost} for the majority of research institutions. Even in that case, repeated occurrences of out-of-distribution data would demand from scratch pre-training or at least fine tuning each time, preventing the scalability of related tasks. 

Another issue strongly interconnected with massive pre-training is the lack of explainability \cite{survey1}, as it is difficult to track what and how a pre-trained model has learned from data. This black box nature poses questions regarding how rare concepts or rare combinations present in data are handled and whether they can be represented with equal success as the more common ones. At the same time, possible data biases, errors and inconsistencies will be reflected in learned representations, with those issues being hard to be captured and resolved beforehand. Despite the surge of Large Language Models (LLMs) \cite{gpt3, chatgpt, gpt4, touvron2023llama, touvron2023llama2, vicuna2023, almazrouei2023falcon, jiang2023mistral, jiang2024mixtral, team2023gemini} , which could serve as ideal implicit knowledge bases \cite{lymperaiou2023contribution} thanks to the massive information they have stored and their emergent abilities \cite{wei2022emergent}, errors and inconsistencies remain pertinent, termed as 'hallucinations' \cite{huang2023survey, zhang2023sirens}.

On the contrary, in the case of explicit knowledge bases, the degree of contribution of the knowledge source can be measured and evaluated, offering valuable transparent insights regarding the role of knowledge.
Such out-of-distribution information is well represented in structured knowledge graphs. Large scale knowledge can complement pre-trained models by extending their understanding to previously unseen concepts, either by substituting the need for extra training, or by enriching existing datasets to achieve more informative, fair and high quality representations, if (re-)training is necessary. Even in that case, pre-training demands can be reduced, achieving similar representation capabilities to larger models pre-trained without additional explicit knowledge. The quality of such representations is somehow controllable, a benefit which can be attributed to the explicit and transparent nature of KGs: issues regarding biases, errors, concept drifts and inconsistencies can be captured and resolved easily, exploiting automatic techniques or manual interventions. In any case, KGs should contain relevant information to the downstream task in order to be beneficial \cite{commonsense-dimensions}.

There are some downsides regarding the usage of explicit knowledge in the form of KGs. First, many KGs may require manual labor for data collection and curation. The same disadvantage also applies on the construction and maintenance of the graph itself. In certain cases, such as in the medical domain, experts are necessary in order to design and construct dedicated KGs. Moreover, there are difficulties regarding alignment and co-operation between different KGs \cite{commonsense-dimensions}, thus sometimes limiting in practice the improvements they offer.

Combining implicit and explicit information can offer advanced capabilities to downstream tasks, as implicit sources can fuse large-scale general knowledge to a model, while explicit sources can fix errors, enrich existing knowledge and increase a model's transparency.

\textbf{Web-crawled knowledge} refers to unstructured knowledge obtained from the web, which is able to combine the benefits present in implicit and explicit knowledge bases. There is no need for labelled data, but also no need for expensive pre-training, which is one major limitation of implicit knowledge. Online sources are readily accessible, while the amount and the content of retrieved knowledge is easily controlled and customized to the task's needs. A questionable part of web-scrapped knowledge is its quality, as it is hard to validate each available web source. Low-quality data may deteriorate the final performance of the model, therefore time and effort has to be invested in techniques that automatically ensure high-quality data. Web knowledge can offer some amount of transparency, as a sentence leading to the final prediction can be tracked, even if reasoning is not as fully explicit, as in cases of structured graphs.

\textbf{Internal} or \textbf{self-knowledge} is a knowledge type that does not rely on any external source, as it is obtained from existing textual and visual data themselves. 
For example, producing a scene graph enables learning more fine-grained representations compared to merely utilizing VL data in their original format \cite{multi4}. Self-knowledge has demonstrated improvements in downstream model performance, especially when detailed disambiguation is necessary, as it enables better associations between existing data. However, self-knowledge does not extend the knowledge a model has already acquired from the data it has been trained on. Furthermore, it is prone to errors associated with the knowledge acquisition process, such as scene graph generation errors. 

\begin{figure*}[h]
    \centering
    \includegraphics[width=\textwidth]{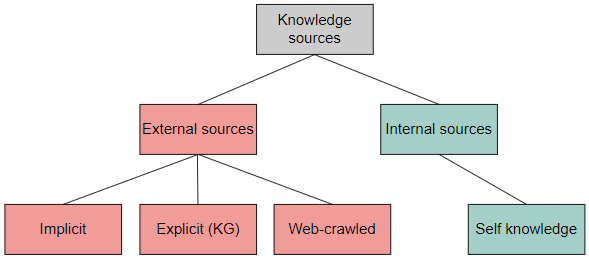}
    \caption{Overview of knowledge sources.}
    \label{fig:sources}
\end{figure*}

An overview of the available types of knowledge sources is provided in Figure \ref{fig:sources}.

\subsection{Widely used Knowledge graphs}
Widely used knowledge graphs in literature are analyzed below.

\textbf{WordNet} \cite{wordnet} is a large-scale lexical database which provides cognitive synonyms, called synsets, for words (nouns, verbs, adjectives and adverbs) of the English language, representing their conceptual and lexical relationships. In total there are 117K WordNet synsets that form a tree-structure through their relationships. Verb and noun synsets are interlinked with transitive hierarchical (is-a) relationships, bringing semantically related synsets close together. Therefore, distance on the WordNet graph provides a measure of concept similarity. There is also a distinction between types and instances, with types (common nouns) expressing more specific meanings of a concept (\textit{cat is a type of animal}), while instances are specific persons, countries and geographic entities (\textit{Rome is an instance of a city}).
The root node of the WordNet tree belongs to the \{entity\} synset, the most generic concept and parent of all other synsets. Traversing from root to leaves leads to more and more specific concepts, with instances always being located at the leaf level. Synsets easily offer word sense disambiguation, as certain words that contain multiple distinct meanings are mapped to a different synset, so that finally all synsets represent different meanings. Part-whole relationships are also present, and parts of a current node can be inherited from parent nodes, but not vice versa. Adjectives are linked via semantic similarity and dissimilarity links, with antonyms and synonyms directly related. 

\textbf{ConceptNet} \cite{conceptnet} is a multilingual commonsense knowledge graph, containing a variety of relationships that express many dimensions of commonsense knowledge. In its current version of ConceptNet 5, it is comprised of $\sim$8 million nodes interconnected by 21 million edges. Specifically, it contains hierarchical relationships (\textit{IsA}), lexical (\textit{DerivedFrom, FormOf}), path-whole (\textit{PartOf, HasA, MadeOf}), similarity (\textit{Synonym, SimilarTo}), dissimilarity/distinctness (\textit{Antonym, DistinctFrom}), spatial (\textit{LocatedNear}), utility (\textit{CapableOf, UsedFor}), creation (\textit{CreatedBy}), quality (\textit{HasProperty, SymbolOf}), desire/goal (\textit{CausesDesire, ObstructedBy}), temporal (\textit{Causes, Entails, HasSubevent}) and relational-other (\textit{RelatedTo, HasContext}) \cite{commonsense-dimensions}. ConceptNet data are aggregated from different sources, with initial versions based on crowdsourcing via the Open Mind Common Sense website. Supplemental sources include other knowledge bases, such as WordNet for lexical information, DBPedia for Wikipedia infoboxes and OpenCyc ontology for high-level commonsense facts, as well as multilingual sources serving similar knowledge content. Additionally, knowledge about intuitive linguistic associations is gathered via so-called 'games with a purpose'.

\textbf{DBPedia} \cite{dbpedia} is a multi-lingual knowledge base that extracts factual structured information from Wikipedia. It covers a variety of domains and evolves together with Wikipedia, thus handling concept drifts. DBPedia also provides SPARQL endpoints to enable queries.

\textbf{Wikidata} \cite{wikidata} is a free collaborative multilingual knowledge base of more than 6.7k relationships and more than 97 million data items, available to everyone for editing. It provides links of entities to their sources and other databases, endorsing the verifiability of contents.

\textbf{WebChild} \cite{webchild} is a large commonsense knowledge base automatically collected from the web. It presents high accuracy of statements, with fine-grained entries involving part-whole, comparative, property, activity and spatial relationships. All entries are disambiguated by mapping on WordNet synsets. In total, WebChild 2.0 \cite{webchild2} contains more than 2 million concepts, interconnected by 18 million relationships.

\textbf{HasPartKB} \cite{haspart} contains part-whole statements extracted from a large generic corpus, covering numerous common terms. Due to the huge possible instances of part-whole relationships in real world, salient parts are preferred, referring to instances that are most probably useful to be stored. All entities are linked with WordNet and Wikipedia.

\textbf{YAGO4} \cite{yago4} is a general purpose knowledge base storing knowledge about people, places, movies, organizations and others. It consists of 2 billion triples and 64 million entities automatically extracted from Wikipedia. WordNet is leveraged for entity and relationhsip disambiguation. YAGO4 actually forms a cleaner and more human-readable version of Wikidata, satisfying logical constrains, therefore allowing automated reasoning on the data.

\textbf{ATOMIC} \cite{atomic} is a commonsense knowledge graph with 1.33M tuples regarding events and entities, incorporating both social and physical senses of everyday experiences. In total, it contains 23 relationship types, with 9 types referring to social interactions, 7 types concerning physical entities and 7 types representing event relationships.

\textbf{Visual Genome} \cite{visualgenome} can be also viewed as a knowledge base, thanks to the scene graph annotations that explicitly showcase entities, relationships and attributes present in a scene, accompanied by the actual visual information of each image. Mappings to WordNet synsets offer disambiguation as well as hierarchical relationships between visual instances. In general, scene graphs act as spatial and visual knowledge bases.

\section{Multimodal Tasks with knowledge}
\label{sec:tasks}
Recent efforts of integrating internal and external knowledge sources in VL tasks are analyzed in this Section, together with datasets and evaluation methods they follow. As mentioned in Section \ref{sec:background}, we can divide existing KVL approaches in \textbf{\textit{single-task}} and \textbf{\textit{multi-task}} models. Starting from \textbf{\textit{single-task}} models, we first present works in the most developed field of \textbf{\textit{discriminative}} KVL tasks, and more specifically \textit{understanding} tasks such as visual question answering (VQA), visual reasoning (VR), visual commonsense reasoning (VCR), followed by \textbf{\textit{generative}} tasks, such as image captioning (IC), visual storytelling (VIST), story visualization (SV), conditional (text-to-image) generation (cIG). Finally, \textbf{\textit{multi-task models}}, targeting more than one downstream tasks at once are analyzed. 

\begin{figure*}[h]
    \centering
    \includegraphics[width=\textwidth]{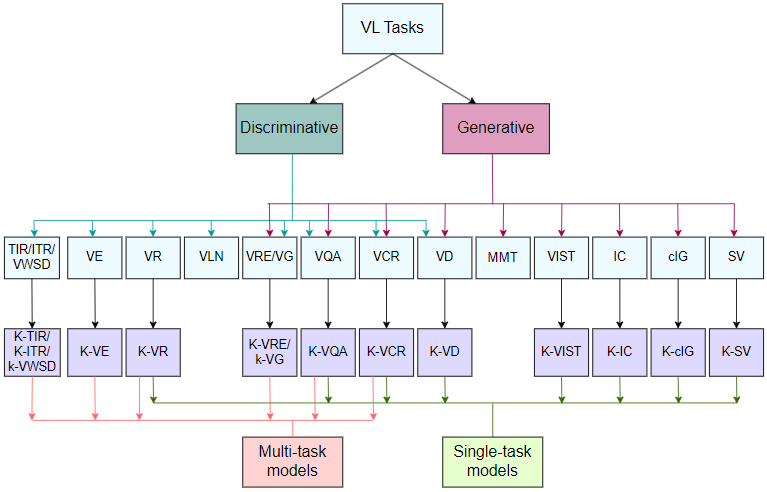}
    \caption{A taxonomy of VL tasks with knowledge}
    \label{fig:task-taxonomy}
\end{figure*}

In Figure \ref{fig:task-taxonomy} an overview and taxonomy of KVL tasks is presented. Some tasks such as visual referring expressions (VRE)\footnote{The inverse VG task (grounding image regions based on linguistic references) can be only discriminative, even though VRE can be either generative or discriminative (generate or retrieve a referring expressions based on an image region).}, visual question answering (VQA), visual commonsense reasoning (VCR) and visual dialog (VD) can be either \textbf{\textit{discriminative}} or \textbf{\textit{generative}}, even though their \textbf{\textit{discriminative}} variants are more widespread, being comparatively easier. \textbf{\textit{Single-task}} models are focusing either on \textit{generative} or \textit{understanding} tasks, while there is no \textbf{\textit{single-task}} model focusing exclusively on a \textit{retrieval} task such as cross-modal retrieval (TIR/ITR) -excluding the Visual Word Sense Disambiguation subtask- and visual referring expressions (VRE). Those two tasks are exclusively tackled -among with others- by \textbf{\textit{multi-task}} models. In the same fashion, visual entailment (VE) is only addressed by \textbf{\textit{multi-task}} models. Finally, a couple of tasks such as visual-language navigation (VLN) and multimodal machine translation (MMT) still lack a knowledge-enhanced counterpart. 

The upcoming Sections are primarily organized starting from \textbf{\textit{single-task}} approaches as per task, followed by \textbf{\textit{multi-task}} approaches in the end. Datasets, methods and evaluation metrics are provided for each independent task.
A more detailed division of methods is achieved based on the knowledge type (external/internal) and the language representation scheme (sequential models/transformers) per approach.

\section{Knowledge in Visual Question Answering (K-VQA)}

\subsection{Datasets}
The following datasets have been used in implementations that have integrated external knowledge sources in order to provide an answer. Many of those, including VQA \cite{vqa, vqav2}, VQA-E \cite{e-vqa}, COCO-QA \cite{coco_qa} are also used in knowledge-free versions of VQA, and external knowledge is not required in order for their questions to be answered.

\textbf{VQA} \cite{vqa} is the first dataset introducing the task of Visual Question Answering. It contains approximately 204k images with diverse and complex scenes, with at least 3 open-ended free-form questions per image and a total of 760k questions in the whole dataset. Many of those questions are commonsense related, such as \textit{Is this a vegeterian pizza?}. For each question 10 ground truth answers are suggested from different annotators, with a total of almost 10 million answers in the dataset. Most answers are short, with the majority of them consisting of a single word. The answers can be evaluated either in open-ended or in multiple-choice settings. Open-ended answers should be validated by 3 annotators agreeing on exactly the same answer for a given question. The multiple-choice scenario regards 18 unique candidate answers per question. Such candidates can be \textit{correct} answers, obtained from the 10 matched answers per question, \textit{plausible} answers, i.e. possibly incorrect answers provided by annotators without viewing the image, \textit{popular} answers, i.e. the most frequently appearing answers in the dataset, and \textit{random} answers sampled from other random questions within the dataset.

\textbf{VQA with Explanations (VQA-E)} \cite{e-vqa} pursues the tractability of the reasoning process leading to an answer. In total, it contains around 108k images, and more than 269k explanations assigned to an equal number of QA pairs. Based on VQAv2 \cite{vqav2}, it automatically constructs explanations with the help of COCO captions \cite{coco}, as they are connected with VQAv2 images. Caption embeddings and question/answer embeddings are coupled, forming pairs of highest cosine similarities, thus assigning captions to images. Resulting explanations are highly diverse, with more than 171k unique instances, although they cannot cover images with subjective questions, such as emotional (\textit{\textbf{Do you} think this pony is cute?}), commonsense (\textit{\textbf{Can you} cross the street?}) or behavioral (\textit{\textbf{Could you} eat all these bananas by yourself?}) ones. Human evaluators assess the quality of explanations, measuring if they are fluent, correct, relevant and complementary to the answer.

\textbf{DAQUAR} \cite{daquar} is a dataset of real world indoor scenes containing fine-grained object categories. Questions and answers related to the images are very rich regarding the objects they express: 573 unique nouns are mentioned within the whole corpus of questions and answers. Questions requiring commonsense knowledge such as \textit{Which object on the table is used for cutting?} are included in DAQUAR, while even spatial questions such as \textit{What is above the desk in front of scissors?} can be benefited from additional knowledge.

\textbf{COCO-QA} \cite{coco_qa} addresses shortcomings of the DAQUAR \cite{daquar} datasets, such as its small size in terms of train/test samples and the limited number of object classes. COCO-QA contains 123,287 images, together with 78,736 train and 38,948 test questions obtained from COCO image descriptions \cite{coco}. Questions are divided in 4 types with varying numbers of questions in each of them: \textit{Object}, \textit{Number}, \textit{Color} and \textit{Location} questions.

\textbf{KB-VQA} \cite{vqa1} has been constructed in order to evaluate VQA models on questions that need visual information as well as external knowledge to explicitly infer the right answer. It includes images from COCO \cite{coco} containing approximately 150 object classes and 100 scene classes, question-answer pairs following pre-defined templates and question labels. The questions involved in KB-VQA are divided in three categories: visual questions can be answered by extracting information from the image (such as \textit{Is there a dog in this
image?}); common-sense questions rely on external knowledge contained in commonsense knowledge bases (\textit{How many road vehicles are in this image?}); finally, KB-knowledge questions require information form Wikipedia or similar sources (\textit{When was the home appliance in this image invented?}). 

\textbf{Factual VQA (FVQA)} \cite{fvqa} is a dataset addressing factual VQA, based on images sampled from COCO \cite{coco} and ImageNet \cite{imagenet} which form three types of visual content (object, scene and action classes), together with structured visual-related knowledge extracted from DBpedia \cite{dbpedia}, ConceptNet \cite{conceptnet} and WebChild \cite{webchild}. All this information is stored in a graph of RDF triplets. Annotators construct questions and answers which require both selected visual content and associated facts. In total, FVQA contains 2,190 images of 326 object classes and 221 scene classes, 5,826 questions of 32 categories, which correspond to 4,216 unique facts. 

\textbf{Knowledge-aware VQA (KVQA)} \cite{vqa5} targets world knowledge-aware VQA by filling the gap of named entities knowledge. It relies on knowledge present in Wikidata \cite{wikidata} KG, resulting in 183k question-answer pairs which involve more than 18k named entities and 24k images. 

\textbf{Outside-knowledge VQA (OK-VQA)} \cite{okvqa} contains more than 14k diverse and difficult questions of 10 mutually exclusive categories which cannot be answered without external knowledge. More than 14k images were sampled and filtered from COCO \cite{coco}. In contrast with previous related works, OK-VQA does not consult a fixed knowledge graph to guide answer prediction, but dynamically recognizes what knowledge is needed, either structured or unstructured.

\textbf{Text-KVQA} \cite{vqa6} is a very large dataset addressing scene-text recognition for knowledge-enabled VQA. It contains images from book covers \cite{book} and movie posters \cite{movie-genre}, as well as Google scraped images of 1000 business brands. All images are evaluated to ensure they contain scene text relevant to the content. Knowledge bases corresponding to each of those 3 scene types were constructed based on Wikidata \cite{wikidata} for business scenes, IMDb \cite{imdb} for movie posters and \citet{book} for book covers. The train/validation/test splits enable zero-shot capabilities, as there is no entity overlap between them. The supporting facts are not tied with their corresponding entities, but instead are dynamically mined from the knowledge bases. 

\textbf{Visual7W+KB} \cite{vqa8} is an extension of the Visual7W \cite{Visual7W} test split which further contains knowledge-based visual questions guided from ConceptNet \cite{conceptnet}. However, the dataset is not tied with a specific knowledge graph, even though ConceptNet is indeed preferred in practice. In total, it consists of 16,850 open-domain question-answer pairs and 8,425 images from Visual Genome \cite{visualgenome}. The questions belong to  7 categories (\textit{what}, \textit{where}, \textit{when}, \textit{who}, \textit{why}, \textit{which} and \textit{how}), while the answers are provided in multiple-choice format.

One of the major challenges in knowledge-enhanced VQA is that questions should encourage exploitation of all participating modalities, therefore data-related weaknesses arise in existing benchmarks. In the meanwhile, information leakage between train and test set answers often promotes guessing rather than reasoning. \textbf{S3VQA} \cite{vqa20} is a dataset designed to address these issues by including questions that require the use of a knowledge graph, along with visual and textual information from the image.

\textbf{Zero-shot Fact VQA (ZS-F-VQA)} \cite{vqa21}  extends F-VQA \cite{fvqa} for zero-shot learning settings. It considers the image-question-answer triples whose answers belong among the 500 most frequent ones. The filtered dataset is split in train (seen) and test (unseen) triples which contain non-overlapping answers. In total, 5 splits of the original F-VQA dataset are performed, yielding on average 2,732 train and 2,760 test triples.  

\textbf{Art QUestion Answering (AQUA)} \cite{aqua} is a visual reasoning dataset for the art domain. There are many challenges tied with analyzing and reasoning over artworks. First, there are different levels of abstraction regarding common objects and entities, as many paintings deviate from realism. Therefore, recognizing objects and reasoning about them is much harder compared to scenes existing in most datasets. Moreover, domain knowledge regarding artists, art movements, historical periods and other cultural influences can only be recognized with the help of a knowledge source. This information also affects the interpretation of a painting. QA pairs are generated automatically based on paintings and descriptions of the SemArt \cite{semart} dataset, which form the knowledge source. In total, AQUA contains more than 69k QA training pairs after cleansing, from which around 29k pairs are visual and 40k pairs are knowledge oriented.

\subsection{Methods}
\subsubsection{Keyword-based explicit KG querying}
First attempts target the construction of a scalable multimodal knowledge base which aims to answer visual queries that require real-world knowledge. Image classes, attributes and actions are extracted from the images, forming logical rules. The knowledge base built upon those rules contains nodes of visual and textual entities, as well as edges of diverse types between the entities. \cite{vqa0} However, this constructed knowledge base remains limited to the visual information present specifically in the SUN \cite{sun} dataset. Most subsequent methods utilize already constructed large knowledge bases, targeting a broader range of concepts, commonsense knowledge and more complex questions to be answered.

Towards this direction, early approaches focus on handling open ended questions regarding contents of a scene with the assistance of provided external knowledge. Attributes extracted from images using a fine-tuned VGG-16 \cite{vggnet} model act as SPARQL queries to knowledge bases such as DBPedia \cite{dbpedia}, and contribute to caption generation. Retrieved knowledge embedded via Doc2Vec \cite{doc2vec}, together with attributes and LSTM-based caption representations are fed in another LSTM model which generates the final answer. \cite{vqa1} An improvement of this version followed in \cite{multi1}, extending the framework to two more datasets, namely DAQURA-ALL \cite{daquar} and its reduced version DAQUAR-REDUCED.

Even from early works in knowledge-enhanced VQA, explainability is recognized as an important aspect in understanding how a model learns from visual content and external knowledge to reach a conclusion. Therefore, Wang et al \cite{vqa2} developed a knowledge-enhanced VQA framework that provides the reasoning path from which the answer is inferred. Objects are detected using Fast-RCNN \cite{fast_rcnn} object detectors trained on ImageNet \cite{imagenet} and MS-COCO \cite{coco}, scene classes are extracted from a VGG-16 \cite{vggnet} pre-trained on MIT-Places \cite{mit_places}, and scene attributes are captured via a VGG-16 pre-trained on ImageNet and fine-tuned on MS-COCO. All those visual concepts form RDF triples and are linked with corresponding DBPedia \cite{dbpedia} entities. Questions are parsed so that key-phrases are extracted and mapped to the knowledge base entities. The same work introduced the KB-VQA dataset. 

Consequent works further proceed towards avoiding SPARQL queries on the knowledge graph, but rather fully utilize embedding representations for fact selection and reasoning to provide an answer.
\subsubsection{Sequential language models for question encoding}
Rather than relying on SPARQL querying from plain keyword extraction, leveraging vector representations of involved modalities forms the foundation for improved performance and state-of-the-art results in knowledge-enhanced VQA. Initially, fact ranking based on embedding similarity metrics paved the path for successful approaches, upon which graph neural network (GNN) reasoning further advanced the contribution of external knowledge and overall performance.

\paragraph{Embedding-based fact retrieval from KG}
Traditional VQA models utilizing RNNs for language encoding, focus on learning a question-answer mapping. Due to the limited and opaque reasoning capabilities of this approach over diverse answers, a more scalable solution that involves learning the mapping between questions and KB-queries using LSTMs was proposed. This new approach is explainable, as the fact connecting a question and an answer reveals the reasoning procedure. \cite{fvqa} 

Projecting question-image pairs and facts on a common embedding space poses advantages over previous approaches, such as extendability to different knowledge bases and error elimination by avoiding explicit querying. Images and questions are embedded using CNNs (for objects, scenes and actions) and an LSTM respectively, and they are projected in a common space using a multi-layer perceptron. Another LSTM is used to retrieve facts from the knowledge base, which are then encoded in GloVe \cite{glove} vectors. The dot similarity between the question-image representation and fact embeddings provides a fact ranking, from which the final answer is inferred. \cite{vqa3}

Narasimhan et al \cite{vqa4}, building upon \cite{vqa3}, argue that considering muiltiple relevant facts instead of a single top-ranked fact at a time leads to better generalization. During the fact retrieval stage, a subset of highly-relevant facts is obtained with the help of LSTM-extracted question embeddings. In the answer prediction stage, each node is represented by concatenating the selected entity representation from the previous stage, visual features from the image and the question embedding. The subgraph formed from all the relevant facts is jointly assessed by a graph convolutional network (GCN) \cite{gcn}, followed by a multi-layer perceptron that decides whether each entity constitutes the final answer or not. 

Based on the KVQA dataset, a memory network (memNet) framework sets the baseline for VQA enhanced with knowledge of named entities. Specifically, entities extracted from the question and the image are used to obtain facts from Wikidata \cite{wikidata} knowledge graph. Retrieved facts together with corresponding entity coordinates from the image are used to produce memory embeddings via a BiLSTM network, and a similar procedure is followed for the question embeddings. Both representations contribute to the final answer, which is defined by a multi-layer perceptron. \cite{vqa5} 

Text present on an image can provide further information towards inferring the correct answer. Extracted text and image areas are fused together with the given question to retrieve relevant facts during the fusion stage, and a multi-relational graph is constructed based on all those components. The text recognition part relies on word proposals assisted by the knowledge graphs accompanying the text-KVQA dataset. Scene proposals were created with the help of Places dataset for scene recognition \cite{places-dataset} and a fine-tuned VGG-16 \cite{vggnet}. A gated graph neural network (GGNN) performs one-hop reasoning on this graph to derive the final answer. \cite{vqa6}

\paragraph{Multimodal graphs}
Unexpected noise in the answer inference process can be attributed to the absence of detailed selection of information during modalities fusion. Considering multiple views of the same image offers a new perspective that is closer to human cognition. Multiple knowledge graphs provide visual, semantic and factual information derived from corresponding images, text and facts respectively, while the visual and the semantic graph can be considered as instances of the factual graph. Intra-modal graph convolutions focus on the most relevant parts of each modality. Consequently, cross-modal knowledge reasoning on the fact graph iteratively aggregates information from the visual and semantic graphs using a recurrent module, and after multi-step reasoning the multimodal knowledge is fused in each entity. The final answer is returned by applying a GCN \cite{gcn} over those entities. This approach offers interpretability by revealing the entity and the modality graph which contributed to the answer. \cite{vqa8} 

A similar approach to \cite{vqa8} utilizes visual, semantic and factual graphs for image representation to eliminate noise during multimodal fusion. The Multi-Modal Heterogeneous Graph Construction stage is responsible for constructing those modality graphs, followed by the Cross-Modal Heterogeneous Graph Reasoning which selects intra-modal knowledge and then performs cross-modal reasoning. Information relevant to the question is extracted from the three graphs via a modality-aware heterogeneous GCN. Cross-modal convolutions define complementary relevant information transmitted from visual and semantic graphs to the fact graph. The final answer is returned after reasoning over aggregated fact information. \cite{vqa10}

A major challenge in knowledge-enhanced multimodal tasks is its supervised nature, as a possible absence of ground truth facts may hinder the inference of a proper answer in several approaches. A local subgraph is constructed based on concepts present in the image and the question, aiming to bridge the gap between question-image context and external knowledge. Those subgraph concepts act as anchor points to a knowledge graph, such as ConceptNet \cite{conceptnet} and Wikidata \cite{wikidata}, enabling the expansion to their immediate neighbors. Moreover, a global subgraph is constructed in a similar fashion for all the candidate answers. In each subgraph, the information of neighboring nodes is aggregated to produce embeddings of the anchor concepts, and their similarity to the query embeddings drive the final answer.   \cite{vqa11}

\paragraph{Multiple feature spaces}
Addressing the zero-shot setting of knowledge-enhanced VQA, a KG can help capturing semantics outside training data. Multiple feature spaces are used for independent alignment between image/question input and KG entities. The semantic space focuses on the linguistic information of the input (image, question) pair, representing a feature space of relationships; the object space acts as a support entity feature space, capturing visual and textual salient features; finally, knowledge space is dedicated to answer representation. \cite{vqa21}

\subsubsection{Transformer-based models}
Transformer based approaches form end-to-end architectures that utilize single-modality or joint representations rather than creating queries to knowledge bases and then injecting the retrieved entities. Thus, we can classify the transformer-based approaches in two categories: the first includes architectures which use transformer architectures for text encoding, while the second utilizes multimodal transformers to jointly encode vision and language.

\paragraph{Transformer architectures for language encoding}
Similarly to \cite{vqa8, vqa10}, the usage of dedicated graphs for different modalities is also followed in \cite{vqa9}, attempting to represent relationships between visual objects and semantic entities present in a scene graph and a knowledge graph respectively. The scene graph is constructed from visual and question embeddings which form the graph nodes and relationships. In the meanwhile, joint image and question embeddings select the most relevant knowledge graph node embeddings to construct the concept graph. Both image-question and knowledge representations are obtained via pre-trained language models for sentence similarity, such as sentence-BERT (SBERT) \cite{sbert} and Universal Sentence Encoder (USE) \cite{use}. The most relevant nodes of both scene and concept graphs are selected via a Graph Attention Network (GAT) \cite{gat}, which decides upon the edge weights with respect to the question. A joint embedding incorporates the question embeddings together with the scene graph and knowledge graph outputs.

Even though explicit and structured knowledge bases constitute the majority of the approaches analyzed so far, implicit and unstructured knowledge can also boost the VQA task. GPT-3 \cite{gpt3} can retrieve knowledge based on text prompts and effectively reason over it in a few-shot manner: no fine-tuning is required and instead only a few examples during inference are provided. Captions are first extracted from images using VinVL \cite{vinVL} to form GPT-3 inputs. Regarding sample selection for the few-shot inference stage, both improving the quality and increasing the number of samples have been explored. The top-n most similar prompt examples comparing to the inference-time question to be answered are defined by CLIP \cite{clip}, thus maximizing sample relevance. On the other hand, multiple queries corresponding to one inference-time example can be used to retrieve answers from GPT-3 using n example prompts each time, and their ensembling results in the final answer. Contrary to most works on knowledge-enhanced VQA, inferring the answer is a generative task and not a discriminative one among pre-selected candidate answers or graph nodes. \cite{vqa15}

Unimodal pre-trained transformers yield better generalization capabilities over multimodal approaches of comparable size when external knowledge is necessary. Language models employed for the knowledge-enhanced VQA task are sufficient even to compensate for the limitations of image captioning models, which often fail to fully capture visual semantics. To this end, a pre-trained image captioning system, in this case the multi-task OSCAR \cite{li2020oscar} transformer, is used to extract linguistic information from an image, while a language model such as BERT, acting as an implicit knowledge source, receives the caption and the question to infer an answer. Moreover, text-only and multimodal approaches have complementary capabilities, therefore their combination can yield even more powerful models. \cite{vqa19}

VIKING is a framework accompanying AQUA dataset \cite{aqua} for VQA on the artistic domain. As questions in AQUA may either be visual or knowledge-oriented, a modality selector first decides the right category by receiving the encoded image and question. Visual-oriented questions do not require external knowledge in order to be answered. For knowledge-oriented questions, a two-stage fact retrieval strategy is followed, pairing the given question with the most relevant painting description, which corresponds to the external knowledge fact needed. The first stage utilizes TF-IDF to rank descriptions according to the question, while in the second stage re-ranking is performed using BERT. Finally, a fine-tuned XLNet \cite{xlnet} model provides the final answer.

\paragraph{Joint multimodal encoding with attention-based fusion}
A slightly different technique is followed in \cite{reasoning1}, where the visual modality is not captioned, but fused with the BERT-embedded question. More specifically, a knowledge graph for artistic VQA is construcyed based on YAGO \cite{yago4} KG and AQUA \cite{aqua} dataset. A Hierarchical-Knowledge Embedding module is responsible of retrieving relevant relationships \textit{r} from the KG, which can form a \textit{(h, r, t)} triple with question related entities serving as the head \textit{h} of the triple, and answer related entities as the tail \textit{t}. A Network-Based Representation Learning module extracts visual and textual features and fuses them together in order to obtain a VL representation. The fusion part first applies local attention per modality, and then global attention on both text and image, where locally 'attended' text features form the query, and visual features the key and value. Query, key and value are inserted in a multi-head attention unit which further promotes the joint representation to consequent layers, until a global representation is obtained. Then, a Knowledge-Based Representation Learning module injects hierarchical-knowledge embeddings to the network-based representation. This representation is inserted into a relational module, which performs meta-training: a representation learned on the training data is transferred to a support set of disjoint labels. Finally, the relational module derives the answer.

\paragraph{Joint multimodal encoding with VL transformers}
ConceptBERT is one of the first attempts towards the end-to-end transformer-based direction, where all modalities are jointly exploited for learning. The first step includes obtaining representations for each individual modality. Visual features are extracted using a pre-trained Faster R-CNN network \cite{faster_rcnn} and BERT \cite{bert} provides the question representation. ConceptNet \cite{conceptnet} acts as the commonsense knowledge source, and is encoded using the ConceptNet embedding \cite{conceptnet_embedding} method, a Graph Convolutional Network \cite{gcn} variant that relies on message passing from node to node in order to obtain the ConceptNet graph representation. Two modules receive the embedded inputs: A vision-language module consisting of two streams in a ViLBERT \cite{lu2019vilbert} fashion, and a concept-language module based on the bidirectional Transformer architecture are proposed to model the interactions between the relevant modalities. Both outputs of these modules are joined to form a concept-vision-language representation, which finally concludes to the answer via a classifier. \cite{vqa7}

Knowledge obtained from the web according to the given question and respective answer can act as a large external implicit knowledge source for the OK-VQA dataset, covering knowledge 'gaps' in several domains without manual human effort. The proposed  weakly-supervised framework consists of two phases: the first one (Retriever) retrieves relevant knowledge, which guides answer prediction in the second stage (Reader). Two different approaches are followed for representing the question-image pair inputs:  either the question and image are encoded using an LXMERT \cite{tan2019lxmert} transformer, resulting in a multimodal representation, or the image caption and the question are encoded via BERT \cite{bert}, leading to an exclusively linguistic representation. The BERT-based linguistic encoding can contribute to both a neural based retriever and a term based \cite{bm25} retriever.
A similarity score defines the relevant knowledge for the neural based retriever, which is further concatenated with the question representation, and consequently with the image using again an LXMERT model.   \cite{vqa12}

The KRISP framework addresses the scenario where essential external knowledge is absent during training, as well as test time. Both implicit and explicit knowledge sources are utilized: Explicit knowledge combines DBPedia \cite{dbpedia}, ConceptNet \cite{conceptnet}, VisualGenome \cite{visualgenome} and hasPart
KB \cite{haspart} in a KG of 36k
edges and 8k nodes after filtering out irrelevant concepts, while pre-training using BERT can offer implicit knowledge. Explicit visual symbols from images are extracted to constrain the KG entities corresponding to image-related concepts, including objects, parts of objects, attributes and places. Likewise, symbols are extracted from question to contribute to the formation of a graph for all explicit symbols. A Relational Graph Convolutional Network (RGCN) is used for graph representation, allowing dedicated processing for different edge types and directions. After reasoning, a symbolic prediction vector is returned. Regarding the implicit information stream, a multimodal BERT (MMBERT) model incorporates visual and textual embeddings to produce an implicit prediction vector. Finally, the top-ranked prediction from both vectors defines the answer. \cite{vqa16}

The presence of scene text can offer valuable information for properly predicting the correct answer. The detected text, among with the image, the relevant knowledge from Google Knowledge Base (GKB) and the question representation are fed into a multimodal transformer, enabling interaction through attention mechanisms between the different modalities. The OCR-extracted text acts as a query to GKB to retrieve candidate entities, which are then disambiguated based on the visual context. External knowledge not only boosts the understanding of scene-text even in unseen instances, but also tackles biases present in training data. \cite{vqa13}

While employing an abundance of knowledge sources is capable of covering more visual topics, a lot of noise may be introduced, as more irrelevant information is retrieved. MAVEx utilizes multi-granular queries to retrieve external knowledge with the purpose of validating and correcting predicted answers among suitable candidates with the help of various knowledge sources. Specifically, a finetuned ViLBERT \cite{lu2019vilbert} model creates a pool of candidate answers, and together with the corresponding question, extracted keywords and phrases from the question and the possible answers are utilized to query external knowledge. Wikipedia, ConceptNet and Google images act as knowledge sources regarding different views of knowledge. Finally, retrieved knowledge instances are matched with the queries to acquire the highest ranked supporting fact, which returns the degree of agreement with respect to candidate answers, guiding decision towards the most trustworthy knowledge source. \cite{vqa14}

Passage retrieval can serve as an answer selection technique to VQA instead of choosing among pre-defined candidate answers. Both sparse and dense retrieval are investigated. For sparse retrieval, given a question and an image, visual clues such as object names and captions are extracted from the image, and BM25 is used to return the k most relevant passages. For dense retrieval, questions and images are jointly encoded in dense vectors using LXMERT \cite{tan2019lxmert}. In any case, retrieved external knowledge can be integrated dynamically from diverse and generic sources, without using a fixed knowledge base. As positive passages are considered the ones containing exactly the ground truth answer. LXMERT  is used to encode the question and the image jointly, while BERT encodes the passage. Finally, dot similarity between them defines the k most relevant passages. \cite{vqa17}

Based on the E-BERT \cite{ebert} strategy of knowledge injection without expensive re-training, LXMERT-encoded linguistic input is modified to incorporate factual knowledge from Wikipedia by aligning Wikipedia2Vec embeddings with BERT WordPiece \cite{wordpiece} vectors. No other change is required within the language encoder's architecture, while the visual encoder remains entirely intact. Only fine-tuning is required for achieving advanced accuracy due to knowledge injection. In the meanwhile, explainability regarding visual and textual modalities is also enhanced. For this purpose, BM-GAE \cite{bm_gae} is employed to extract visual and token explanations, helping the identification of parts to which knowledge injection was helpful. \cite{vqa18}

S3, presented with the S3VQA dataset, targets to answer visual question based on all the participating modalities simultaneously. Entity spans from the question are selected to be matched with objects of scene graphs corresponding to images. This matching can be often guided by external knowledge sources, which enable answering more complex questions that require multi-hop reasoning. BERT identifies those appropriate question spans, while object detectors propose the objects that most likely fill the spans. Wordnet \cite{wordnet} synsets are mapped to the objects, and their hierarchical positions are represented via structural embedding methods. Finally, Google search is used to retrieved the top results of the enriched question representation. Alternatively, the answer can be provided via classification of possible candidate answers. \cite{vqa20}

\subsection{Evaluation}
\textbf{Classification/Ranking metrics} are widely used in K-VQA, following the paradigm of knowledge-free VQA, with most works relying on the \textbf{top-1 accuracy} metric for comparison. Accuracy is further decomposed to explain the contribution of subcomponents in many works: object, counting, color, and location accuracies, as well as accuracy per question type are reported \cite{vqa1, vqa2}. Other accuracy reportings include performance as per selected knowledge source, per visual concept and per answer source (image or knowledge) \cite{fvqa}; the individual accuracies of involved stages of the reasoning process, which together contribute to answer prediction \cite{vqa3, vqa4}; accuracies as per question category \cite{vqa5, vqa8, vqa9, vqa7}. Moreover, \textbf{precision@k} and \textbf{recall@k} have appeared in fewer works, as well as ranking metrics such as \textbf{MRR}. Some early works \cite{vqa1, fvqa} utilize \textbf{WUPS} with typically used thresholds of 0.0 and 0.9. 

\textbf{Human evaluation} provides further insights in comparison to to solid accuracy-based scores that fail to fully describe the success or the shortcomings of a metric. Therefore, many authors propose human evaluation experiments to grade the model's response comparing to human perception, and count the number of agreement instances over all results \cite{vqa2, fvqa, vqa4}.

\textbf{Evaluation of reasoning paths} is followed by employing human judgement, as being one of the most trustworthy indicators \cite{vqa2}. Due to the transparent reasoning process, failure cases can be traced down, revealing the exact stage where the prediction deviated from the intended one. Thus, shortcomings can be attributed to architectural choices, encoding techniques, or even incorrect data annotations. Other metrics regarding explainable reasoning include \textbf{top-k fact retrieval accuracy} for different k values, a crucial step for returning the correct answer in several approaches \cite{fvqa, vqa3, vqa4, vqa10}. \textbf{Fact recall} can also assess the fraction of relevant facts retrieved for a given question \cite{vqa6}.

Generally, benchmarking knowledge-enhanced VQA approaches is not trivial. The plethora of combinations between available knowledge-based datasets and external knowledge sources is rather large compared to the number of available implementations in the field. Additionally, various choice of metrics in literature makes comparisons of model performance even harder. Only recently implementations started becoming more consistent, focusing on evaluating results with plain accuracy and leveraging OK-VQA as a widely used dataset.

\section{Knowledge in Visual Word Sense Disambiguation (K-VWSD)}
\subsection{Datasets}
\textbf{VWSD} \cite{raganato-etal-2023-semeval} is a recently introduced dataset where an ambiguous phrase (e.g. \textit{Andromeda tree}) comprised of an ambiguous target word (\textit{Andromeda}) and limited context (\textit{tree}) is tasked to retrieve the most suitable image among a set of candidates, in which distractors are present. 

\subsection{Methods}
\subsubsection{Transformer-based models}
\paragraph{Explicit knowledge}
Knowledge-enhanced approaches are capable of expanding and specifying the ambiguous target with more context, therefore adopted in all top-performing VWSD systems. A multimodal knowledge base containing words, senses (for ambiguous targets) and images provides appropriate word-sense-image features to a fine-grained image-text matching module in a contrastive fashion, leveraging WordNet to obtain the different word senses; this approach results in top results among all participating languages. \cite{yang-etal-2023-tam} Knowledge-based sense inventories crafted using dictionaries provide the necessary options to a biencoder architecture which decides on the most appropriate sense, which is consequently provided to a pre-trained VL transformer. \cite{zhang-etal-2023-srcb} Phrase enrichment from Wikipedia and dictionaries facilitate text-image retrieval and the integration of multimodal features in a learning-to-rank module. \cite{dadas-2023-opi-semeval} More related approaches using the lexical knowledge of dictionaries and relevant knowledge bases \cite{grbowiec-2023-opi, schneider-biemann-2023-lt, zhang-etal-2023-gpl} achieve comparably lower results.

Opting for language models as knowledge bases, \citet{ghahroodi-etal-2023-sut} leverage ChatGPT to decide upon the proper sense based on a pre-defined prompt template; nevertheless, this approach scores lower than the ones that step upon deterministic knowledge bases. On the contrary, optimal prompt template selection is capable of enriching the ambiguous phrase, leading to top-performing LLM-enhanced systems for VWSD, comparable to the ones built upon explicit knowledge bases. \cite{emnlp} Moreover, thorough experimentation with prompting techniques, such as zero-shot and few-shot prompting establish LLMs as a suitable knowledge base for phrase disambiguation, and thus for boosting VWSD results. \cite{iswc}

\subsection{Evaluation}
\textbf{Ranking metrics} naturally fit the K-VWSD task; specifically, accuracy@1 decides if the top-ranked image correlates with the ground truth one, while MRR evaluates the ranking quality by also considering the rank position of the ground-truth image. \cite{raganato-etal-2023-semeval}

\section{Knowledge in Visual Reasoning (K-VR)}
\subsection{Datasets}
\textbf{High-order Visual Question Reasoning (HVQR)} \cite{hvqr} is a knowledge-based dataset endorsing interpretable visual reasoning using commonsense knowledge. Given an image and a question, an answer is inferred, as well as a reasoning path as explanation. Even though this is similar to rationales used in knowledge-free datasets for VCR \cite{vcr}, in fact the format of HVQR explanations differ. Instead of textual rationales, rules for the whole reasoning path are returned, combining visual and knowledge-oriented triples, derived from the scene graph and the commonsense knowledge graph respectively. HVQR contains questions that require multi-step reasoning to infer an answer. Moreover, each knowledge triplet appears only once per question, in order to avoid frequency-based biases. An evaluation scheme validates each step of the reasoning process based on the commonsense and scene graphs provided. More than 157k QA pairs comprise the dataset,  together with approximately 32k images and corresponding scene graphs from Visual Genome \cite{visualgenome}. Based on the reasoning steps required for the answer, first-order and second-order questions can be recognized, corresponding to 68,448 and 88,753 questions
respectively. Another split defines 87k KB-related questions and 70k KB-not-related questions. Additionally, 193,449 facts from WebChild \cite{webchild}, ConceptNet \cite{conceptnet}, and DBpedia \cite{dbpedia} formulate the knowledge base. Scene graphs per image are combined with related entities from the knowledge base, constituting image-specific knowledge graphs.

\textbf{Compositional Language and Elementary Visual Reasoning - CLEVR} \cite{clevr} is a synthetic dataset of 3D objects which contain annotations regarding their position and attributes. Those attributes describe the size \textit{(small, large)}, color \textit{(red, brown, yellow, green, blue, cyan, purple, gray)}, shape \textit{(cube, cylinder, sphere)} and material \textit{(rubber, metallic)} of each object. Positions can belong in 4 types, namely \textit{left}, \textit{right}, \textit{behind}, \textit{in front}. Highly compositional questions form 5 question categories: \textit{Exist}, \textit{Count}, \textit{Compare Integer (equal, less, greater)}, \textit{Query Attribute (size, color, material, shape)} and \textit{Compare Attribute (size, color, material, shape)}. Moreover, CLEVR contains 90 question families following different program templates, as well as text templates, so that natural language questions can be derived. The questions are translated in natural language by filling the template with template parameters. CLEVR is not connected to some external knowledge, although due to the limited semantics and the nature of the task they target 

\textbf{CLEVR CoGenT} is a benchmark derived from CLEVR \cite{clevr} that assesses the ability to capture unseen combinations of attributes during testing, thus showcasing a model's generalization capabilities. 

Other datasets associated with visual reasoning tasks that have not yet participated in knowledge-enhanced endeavors are the \textbf{Kandinsky Patterns}, \textbf{NLVR}, \textbf{NLVR2}, \textbf{Winoground}, \textbf{WinoGAViL}.
\textbf{Kandinsky Patterns} \cite{kandinsky} is another synthetic dataset containing geometrical shapes in the 2D space, characterized by their shape, color and position. A sequence of images within the Kandinsky Patterns dataset is formed based on descriptions and rules regarding the participating objects, i.e. \textit{the Kandinsky Figure has two pairs of objects with the same shape, in one pair the objects have the same color, in the other pair different colors, two pairs are always disjunct, i.e. they don't share objects}. \textbf{NLVR} \cite{nlvr} contains statements such as \textit{There are two towers with the same height but their base is not the same in color}, again requiring reasoning over image sequences that contain 2D geometrical shapes of certain colors, shapes and positions. \textbf{NLVR2} \cite{nlvr2} expands the complexity of images while maintaining the problem formulation introduced in NLVR. Compositional reasoning is evaluated in the recently introduced \textbf{Winoground} dataset \cite{thrush2022winoground}, which contains pairs of images and pairs of captions and requests proper image-text matching; the challenge is that captions contain the same words and word order defines the correct image matching e.g. \textit{some plants surrounding a lightbulb} vs \textit{a lightbulb surrounding some plants}. Multiple reasoning skills are evaluated in the \textbf{WinoGAViL} dataset \cite{bitton2022winogavil}, where a human player sets a linguistic association cue, while an AI rival attempts to select the most appropriate images to the cue. A more thorough investigation of visual reasoning datasets is provided in \citet{reasoning-datasets}.

\subsection{Methods}
\subsubsection{Sequential language models}
\paragraph{External knowledge}
KM-net (Knowledge-routed Modular Network) is introduced in the same work with the HVQR dataset \cite{hvqr}, addressing multi-step (compositional) reasoning using visual and commonsense knowledge. Each question is decomposed into consecutive subqueries via LSTM encoder-decoder schemes, passed to a visual reasoning module and a commonsense reasoning module to extract different types of knowledge accordingly. The subqueries form a query layout, i.e. a tree structure revealing the relationships of subqueries, with leaf nodes belonging to distinct words of the queries. A bottom-up attention R-CNN provides visual features for the image. The subqueries are processed sequentially starting from the most specific ones, driven by the KM-net reasoning module. First, the knowledge reasoning module receives question entities from the query layout and returns the most probable candidate entities from the knowledge base. Then, the visual reasoning module receives entities from the scene graph, together with the candidate entities of the knowledge module and fuses the candidate entities, image features and query embedding to derive the answer.

\paragraph{Internal knowledge}
Self-knowledge includes the usage or construction of a scene graph based on the detected objects, relationships and attributes. At the same time, the question can be parsed in a structured program via an LSTM, producing subqueries that form a tree structure. Given those two graph representations, an encoding is derived for the query. Node attention and edge attention are calculated based on the query embedding. Combining a node attention vector and an edge attention matrix, new objects can inferred due to the graph structure; basically starting from the attended node vector and traversing over an attended edge, a new node vector will be provided. The same procedure can be followed for all subqueries, respecting the structure of the query tree. Logically relating subqueries, results in logical operations (such as \textbf{\textit{and}}, \textbf{\textit{or}}, \textbf{\textit{not}}) over attended scene graphs. Finally, based on the question type and the final scene graph, the answer can be provided. \cite{reasoning3}
 
\subsection{Evaluation}
\textbf{Classification metrics} are commonly used for benchmarking, with \textbf{answer accuracy} providing a general measure of performance \cite{hvqr, reasoning3}.
Compositional commonsense reasoning heavily relies on the evaluation of reasoning paths that provide the final answer \cite{hvqr}. The accuracy score is further decomposed to present \textit{KB-related} and \textit{KB-not-related} accuracies depending of the need for external knowledge; those can be broken down to \textit{first-order} and \textit{second-order} accuracies, regarding the number of reasoning steps required; finally, a more fine grained categorization provides
\textit{question-type accuracy}, based on the template the query components follow.

\textbf{Ranking metrics} such as \textbf{average recall} are used to evaluate the retrieval success of supporting facts for explanations. Average recall is further decomposed to \textit{KB-related} and \textit{KB-not-related} fact recall.

\section{Knowledge in Visual Commonsense Reasoning (K-VCR)}
Various external knowledge sources rich in information can provide insights of unseen concepts that humans would effortlessly infer from the information provided in a scene. This missing commonsense knowledge is able to guide answer explanation towards the right rationale, revealing if a more accurate reasoning process is followed by VCR models when knowledge is added.
\subsection{Datasets}
\textbf{Visual Commonsense Reasoning (VCR)} \cite{vcr} is the dataset which introduced the task and serves both knowledge-free and knowledge-enhanced versions of VCR. It contains 110k unique images from movie scenes, 290k multiple choice challenging questions, with 290k correct answers and rationales. Images contain annotations which are anchored over questions, answers and rationales. The technique of adversarial matching is chosen for the answers in order to minimize biases; each correct answer appears four times in the whole dataset, once as a positive answer and three times as negative answer. Therefore, a VCR model will not favor more frequently appearing answers which would endorse guessing rather than reasoning. The questions are classified in non-mutually exclusive categories according to their purpose, with categories being Explanations (\textit{Why is [person11] wearing
sunglasses inside?}), Activity (\textit{What are [person1] and
[person2] doing?}), Temporal (\textit{What will [person6] do after
unpacking the groceries?}), Mental (\textit{What is [person3] thinking while
[person5] shakes his hand?}), Role (\textit{What is [person1]’s relation to
[person4]?}), Scene (\textit{Where is [person1] now?}), Hypothetical (\textit{What would happen if [person3] fell asleep?}). This dataset originally is knowledge-free, therefore not necessarily requiring external knowledge, nor is it associated with any knowledge base. Nevertheless, the questions existing in the various VCR  question categories can be greatly benefited from the introduction of external knowledge sources which can explicitly incorporate senses like the ones described in Section \ref{sec:knowledge}.

\textbf{Visual Commonsense Graphs (VCG)} \cite{visualcomet} is a large-scale dataset that provides information regarding temporal commonsense relationships, such as \textit{what may have happened before}, \textit{what may happen in the near future} and \textit{what are the intents of the people present} based on static images. In total, it contains more than 59k images and more than 139k textual descriptions of events at present. Additionally, around 295k intents at present, as well as more than 584k events before and 586k events after complete the dataset, resulting in more than 1,4 million commonsense inferences. People and locations appearing in the images are grounded with their mentions in the textual descriptions. 

\subsection{Methods}

\subsubsection{Transformer-based models}
\paragraph{External knowledge}
Some of the first knowledge-enhanced transformer-based attempts step upon BERT \cite{bert} to introduce knowledge-vision-language (KVL) learning as an instance of multimodal learning. In the KVL-BERT architecture \cite{vcr6}, ConceptNet \cite{conceptnet} is leveraged to enrich sentences with relevant commonsense information. The knowledge-enriched linguistic input is inserted in a BERT-like multimodal transformer. The preservation of semantic structure is achieved by using relative position embeddings. However, injected information should be only visible to their corresponding textual entities of the sentence and not to other tokens or visual features, a need that is satisfied via a 'weakening' visible matrix. Moreover, it is possible that different enriched textual tokens in the sentence share the same relative position embeddings, which would make unrelated tokens obtain high self-attention scores, implying that they are related. This contradiction is resolved by imposing a mask-self-attention mechanism via the visible matrix, with the purpose to restrict the area a token can attend. After those treatments, the input is in a form suitable to be fed in a VL transformer, in this case VL-BERT \cite{su2020vlbert}. It was observed that KVL-BERT outperforms its multi-task baselines, as well as models dedicated to the VCR task, even though it cannot trespass the performance of knowledge-free VL transformers that invest on additional pre-training.

A somehow different strategy is employed in the case of Vision–Language–Knowledge Co-Embedding (ViLaKC) \cite{vcr2}: the three modalities are first embedded independently and afterwards are fused together. Initially, a Knowledge Extraction Module (KEM) retrieves relevant knowledge from ConceptNet based on concepts appearing on the image, question and candidate answers. The encoding of modalities is performed in the two-stage VLKEM module: first, the independent modality encoding embeds images using ResNet \cite{resnet}, language using BERT \cite{bert} and knowledge using GCN \cite{gcn}. The second stage consists of the co-embedding sub-module which aligns and integrates the three vectors via a multi-head self-attention mechanism. The co-embedder is pre-trained in two phases, the first being task-agnostic, such as in several VL transformer models, and the second task-specific, utilizing significantly less data ($\sim$200k samples) coming from all three modalities. The task-specific pre-training stage introduces novel pre-training tasks, such as masked language modeling with image and knowledge (MLMIK), masked object classification with text and knowledge (MOCTK), and vision-language-knowledge matching (VLKM), in order to enforce co-learning. This joint embedding is then inserted in an answer determination module (ADM) consisting of a fully connected layer followed by a softmax.

The CKRM framework \cite{vcr8} consists of two stages, the first used for knowledge retrieval and the second one for reasoning. SWAG \cite{swag}, a dataset containing pairs of events which describe a situation (context) and possible endings, serves as the commonsense knowledge source, aiming to transfer knowledge regarding everyday situations to the target task of VCR. A source and a task encoder are responsible of receiving (\textit{context, ending}) pairs and (\textit{question, answer}) pairs respectively to perform knowledge transfer in different granularity layers. The encoders first use BERT \cite{bert} followed by a BiLSTM structure to model temporal interactions of words. Cell-level knowledge transfer refers to the most fine-grained information fusion from source to target task, with layer-level and attention-level knowledge corresponding to coarser aspects of information. This strategy offers acquiring knowledge from various perspectives for a more enriched representation. The knowledge-based reasoning module incorporates the multi-level knowledge from the previous stage together with visual features in the Knowledge-enriched visual attention module. Finally, a reasoning composition module combines all aspects of knowledge derived from the multi-level transfer procedure and the enriched visual representations to derive the answer. 

\subsection{Evaluation}
\textbf{Classification metrics}, especially \textbf{classification accuracy} is employed for evaluating K-VCR results that follow the multiple-choice format for answers (\textit{A}) and rationales (\textit{R}). Accuracy is decomposed by evaluating independently each of the following aspects: 
\begin{enumerate}
    \item $Q \longrightarrow A$: given a question Q, choose as A one of the 4 available answers and compare if it matches the real answer or not.
    \item $QA \longrightarrow R$: given a question Q and the correct answer A, select as R one out of the 4 rationales and compare if it matches the real rationale or not.
    \item $Q \longrightarrow AR$: given a question Q select as A one of the 4 answers, and depending on selected A choose one of the 4 rationales. The result is regarded to be correct if and only if both right A and R are chosen.
\end{enumerate}

\section{Knowledge in Image Captioning (K-IC)}
\subsection{Datasets}
There are no dedicated knowledge-enhanced or knowledge demanding datasets for K-IC. Knowledge-enhanced models are currently using 
\textbf{COCO captions} \cite{coco} as described in Section \ref{sec:pretraining_datasets}. Moreover, 
\textbf{Flickr30k} \cite{flickr}, a dataset containing 31,783 scene images accompanied by 5 human-annotated sentences each is widely employed for K-IC. 
\subsection{Methods}
\subsubsection{Sequential language models}
\paragraph{External knowledge}
First attempts for knowledge-enhanced image captioning propose the extension of existing implementations by injecting commonsense knowledge from external sources. Specifically, the backbone image captioning architecture extracts visual features from images via a CNN, which are then inserted to an LSTM to generate a knowledge-free answer. To enhance this baseline with knowledge, objects extracted from the image are used as queries to ConceptNet \cite{conceptnet}. Related ConceptNet entities, either regarding individual objects (direct terms) or the remaining image areas (indirect terms), are fed to a pre-trained LSTM which provides semantic representations for each of those two points of view. Then, visual features, direct term representations and indirect term representations are concatenated to form the initial state of another LSTM model, which finally generates the knowledge-enhanced caption \cite{caption1}.

Both visual and commonsense knowledge for image captioning are used in \citet{caption2}. The first step includes dense region sampling from images in order to acquire visual and knowledge mappings. Dense visual feature extraction includes the definition of candidate regions, which are clustered together to provide a more concrete representation: the cluster center points for each dense region cluster serve as the corresponding visual feature. Consequently, the knowledge mapping receives visual features and knowledge embedding vectors from Visual Genome \cite{visualgenome} and returns a knowledge-related representation per region cluster. Both visual and knowledge embeddings resulting from the two mapping procedures are concatenated and then inserted to a commonsense reasoning module. This module projects the two inputs on the same semantic space, from which a semantic graph is constructed under the guidance of commonsense knowledge. In the relational reasoning module, a GCN \cite{gcn} operates on the semantic graph to obtain relation-aware node features. Finally, a LSTM receiving the knowledge-aware node embeddings as inputs generates the caption. 

Inferring words not appearing in the image remains a challenge in image captioning, as there is no guidance regarding how those unseen words should be inferred to be used in captions. Such unmatched elements can be solved with internal self-knowledge based on more fine-grained alignments between individual words and image regions, which is achieved by attention mechanisms, and external commonsense knowledge to capture implicit information that cannot be derived from the existing data. Objects detected on the image are used to retrieve knowledge from ConceptNet \cite{conceptnet}. Region features extracted from the image via a region proposal network and word-level attention on the sentence part co-operate towards attending to the most salient features of the image. This visual attention guided by language attention, together with the corresponding word embedding are inserted in a LSTM, which feedbacks each previous hidden state to update the word-level attention signal that contributes to the visual attention in every round. The external knowledge is incorporated in a later stage, when the answer is generated; therefore, it can tune the probabilities of LSTM-generated words to be added in the sentence towards more meaningful results. A reinforcement learning training strategy is followed by setting the LSTM as an agent, the words and visual features as the environment, and the generation of the best next word from the captioning model as the policy. \cite{caption3}

Even though local information is well-represented based on detected objects, image captioning is generally not interpretable and therefore not explicitly controllable. An external knowledge source can help in grounding detected objects with semantic entities from the graph, which in turn provides enriched semantic labels for the objects present in the image. In order to control objects appearing in the caption, an attention-based human-interpretable mask is introduced, which assists in diverse caption generation. This masked can be dynamically tuned by a human to influence the resulting caption. \cite{caption5}

Off-the-self object detectors have served several image captioning architectures. However, some tough situations, such as very small objects, occlusion or rare object classes can result in error propagation and negatively impact all consequent components until the final caption generation. Commonsense constrains and semantic correlations extracted from Visual Genome \cite{visualgenome} can act as priors to guide a more accurate representation. A semantic graph is constructed upon extracted image regions, allowing GCN-based \cite{gcn} reasoning. Specifically, visual semantics such as objects, attributes, relationships are captured by extracting candidate region proposals. CNN-based region features satisfy object and attribute representations, while features from regions union areas provide relationship representation. Visual features are projected on the same high-level semantic space as the knowledge embedding derived from Visual Genome. Therefore, knowledge-enhanced visual triplets are formed, respecting rules imposed by knowledge. The semantic graph is built upon those triples. Then, relational reasoning is performed on the semantic graph using a GCN, the output of which is inserted in the LSTM module that generates the answer. \cite{caption6}

\paragraph{Internal knowledge}
Visiolinguistic priors are naturally connected with describing images, in the sense that humans logically infer unseen entities given a partial description of a visual situation. Obtaining such priors from existing images and captions is a way of 'creating' knowledge and facilitate reasoning of image captioning models without adding external sources.

Scene graph generation is a widely used technique for self-augmentation of information present in the dataset. Both images and text need to be represented within graph structures to bridge the two modalities. The  Scene Graph Auto-Encoder (SGAE) framework utilizes this graph conversion to instill language priors into the encoder-decoder image captioning structure. More specifically, a learnable dictionary maps the relationships between a sentence and its corresponding scene graph iteratively, reconstructing the initial text from the generated graph in each round. For scene graph generation from text, a pre-trained scene graph parser is utilized, while for the reverse procedure, a trainable RNN decoder converts the dictionary back to text. During this procedure, the dictionary achieves to capture the necessary language prior to be transferred for captioning. The learned dictionary is then inserted to the image-involving pipeline: a scene graph parser converts the image to a scene graph, which is consequently passed to the dictionary encoded by a GCN \cite{gcn}. Finally, the decoding of the dictionary provides the final caption.  \cite{caption0}

Attention mechanisms are able to identify such structured visiolinguistic priors and highlight connections between text and images, therefore augmenting image captioning implementations. Conditional Latent Topic Attention (CLTA) in combination with sentence priors are able to fuse the model with prior knowledge without the need for constructing scene graphs. Latent topic models are able to recognize semantically significant topics which are driving attention mechanisms to capture local and global dependencies in images. Thus, salient visual features emerge through words, and also more candidate salient regions are discovered and re-weighted accordingly, if they are associated with a topic contributing to an existing salient region. CLTA implements this re-weighting procedure to construct a context vector. Moreover, a sentence autoencoder acting as the sentence prior encourages the extraction of more context information and enhances generalization. Both the context vector and the sentence prior are inserted in an LSTM that generates the answer.  \cite{caption4}

\subsubsection{Transformer-based models}
Recent knowledge-enhanced image captioning models are implemented based on Transformers as an expected substitution of sequential models. 

\paragraph{External knowledge}
Named entities and event knowledge have not been studied in previous image captioning works. This type of information is widely available in news articles, with raw sources being too complicated for language models to infer the right semantics. Special datasets are crafted for this purpose, providing an appropriate form of information for named entity/event-aware image captioning. The heart of the proposed method is the cross-modal entity matching, which incorporates information from various sources. Sub-graphs are extracted from the image and the article text descriptions forming structured representations for the input. The nodes of the text sub-graph belong to named entities, and the edges to their in-between relationships, while the image sub-graph is more generic, by representing objects present in the image. The two sub-graphs are linked via similarity between image sub-graph objects and text sub-graph named entities in the cross-modal entity matching module. This module is trained with the help of multimodal external knowledge from Wikipedia. As a result, a multimodal knowledge graph is produced containing visual, textual and knowledge information. Embedding representations are obtained for each modality: a GAT module \cite{gat} produces a multimodal knowledge graph embedding, RoBERTa \cite{roberta} encodes news captions and image features are derived from a pre-trained ResNet-152 \cite{resnet}. An entity-aware captioning model receives the visual, textual and multimodal knowledge graph representations, feeding them to a Transformer \cite{transformer} decoder to produce the caption. \cite{caption10}

The BART transformer \cite{bart} can provide further advancements towards the refined task of \textit{Visual Commonsense Generation (VCG)} \cite{visualcomet} lying on the intersection of the generative image captioning task and the non-generative visual commonsense reasoning task. To this end, knowledge-enhanced Multimodal
BART (KM-BART) was developed, able to incorporate both visual and linguistic information with the help of modality and task-relevant tokens in the transformer input. More specifically, task-relevant tokens are added in the beginning of the input sequence denoting the task type. For example, for VCG $<$before$>$, $<$after$>$, or $<$intent$>$ tokens, representing temporal sequence of events (what happened before, what may happen next) and intents of people present in the image.  Furthermore, the pre-training task of Knowledge-based Commonsense Generation (KCG) fuses commonsense knowledge from structured sources early in the pipeline, actually achieving in implicitly integrating explicit knowledge. Also Attribution Prediction (AP) and Relation Prediction (RP) pre-training tasks are used for the first time in knowledge-enhanced VL learning. COMET \cite{comet} is a transformer model trained on knowledge bases such as ATOMIC \cite{atomic} and ConceptNet \cite{conceptnet} that generates commonsense descriptions, and acts as a knowledge source for KM-BART. Two possible settings are examined for KM-BART, one containing the image and the event description (i.e. some textual information about the image that provides context of the depicted situation) and a harder one that omits the event description. \cite{caption8}

\paragraph{Internal knowledge}
Transformer-based captioning poses some challenges, one of those attributed to the AR training procedure which is based on the maximum likelihood estimation (MLE). The main issue stemming from MLE is that when the generated sequence does not match the ground truth one, there is no discrimination between different 'failed' predictions. Therefore, words that are totally unrelated to the ground truth match are treated the same as semantically similar words. For this reason, a KL divergence term is added to weight semantic relationships between generated words, with respect to their ground truth match. Moreover, a knowledge graph is used to enrich the transformer input embeddings, infusing contextual information from neighboring entities in the graph. This knowledge graph is constructed from the linguistic information itself, by leveraging cosine similarity between embedded words to position them within a vector space. The original Transformer \cite{transformer} architecture is used for the task, with image features word embeddings representing the visual modality. \cite{caption9}

\subsection{Evaluation}
\textbf{Language metrics} such as \textbf{BLEU}, \textbf{ROUGE}, \textbf{METEOR}, \textbf{CIDEr} are used for evaluation as in most language generation tasks \cite{caption1, caption2, caption4, caption3}. SPICE is also used in \cite{caption5, caption9}.

\textbf{Human Evaluation} can qualitatively evaluate generated sentences. The human evaluation experiment in \citet{caption0} compares the quality of generated captions from different models according to the perception of 30 evaluators. Even though such an experiment is rather subjective, it indicates the importance of language priors. In \citet{caption10} human preference is measured comparing with the previous best-performer in the \textit{before, after, intent} generated sentences.

\section{Knowledge in Visual Dialog (K-VD)}
\subsection{Datasets}
\textbf{VisDial} \cite{visdial} is a dataset used in both knowledge-free and knowledge-enhanced versions of VD. It consists of 133k dialogs  and an equal number of images from COCO, with train and validation splits (125k dialogs) assigning 10-round dialogs -QA pairs- per image. In the test split (8k dialogs), random rounds are paired with each image. Some important aspects of this dataset is the presence of coreferences, endorsing the coherence of the conversation in linguistic level, and temporal continuity of topics, which supports the preservation and consistency of semantic meaning across the dialogs. The questions mostly follow a concrete and rather exploratory pattern: starting from asking about entities involved in COCO captions, then diving into details, trying to define a categorization of the whole scene or the most appropriate setting description, questioning about the weather of the scene, exploring key semantics not mentioned previously and finally validating and expanding the understanding of elements provided in the answers.

\textbf{VisDialCK} \cite{visualdialog1}  is an extension of VisDial containing 940 history-required and commonsense-required dialogs.  

\subsection{Methods}
\subsubsection{Transformer-based models}
\paragraph{Internal knowledge}
Only one knowledge-enhanced implementation for visual dialog has been introduced, inspired from the fact that commonsense related questions are typically ignored. A visual dialog model requires two necessary inputs: an image and the dialog history. Visual graphs have assisted the task by providing object relationships explicitly, even though this knowledge is not adequate for commonsense inferences.  The integration of commonsense knowledge can be well-represented with graph-level facts and sentence-level facts. Then, facts from a commonsense knowledge graph such as ConceptNet \cite{conceptnet} can be extracted based on the calculation of cosine similarity between their word embedding representation compared to embedding representations of the words in the sentences and the detected objects. Those graph level facts can complement entities from the visual graph. Therefore, an enriched vision-fact graph can be produced after individual graphs are purified by removing redundant information. The sentence-level facts are derived from the dialog sentences in the form of \textit{(subject, relation, object)} triples, forming a graph structure. Similarly to the visual stream, the sentence graph is cleaned and enriched with commonsense knowledge. Finally, a transformer-based fusion module receives the enriched graphs, as well as the question embedding to provide the answer, exploiting a generative and a discriminative decoder.  \cite{visualdialog1}

\subsection{Evaluation}
\textbf{Ranking metrics} such as NDCG, MRR, R@1, R@5, R@10, Mean position provide the quality of answer retrieval for visual dialog for both \textbf{\textit{generative}} and \textbf{\textit{discriminative}} answer prediction. \cite{visualdialog1}

\textbf{Human evaluation} is used in the \textbf{\textit{generative}} setting of \cite{visualdialog1}. Specifically, two metrics are provided: the first one indicates the percentage of responses passing the Turing test, thus providing the amount of generated sentences that could be perceived as human-written; the second metric measures the number of generated responses that are perceived as of equal or better quality compared to specific human responses as baselines.

\section{Knowledge in Visual Storytelling (K-VIST)}
Visual Storytelling presents many situations where hypothetical concepts can be driven from commonsense and temporal reasoning. Unseen events can enrich or even be necessary for appropriate and coherent textual stories. For example, some sequential inferences were presented in the event/temporal knowledge analysis (in Section \ref{temporal}), providing knowledge such as \textit{the boy dropped a glass of water and then the glass broke}. Not all concepts mentioned in this sentence may be explicitly apparent on a frame of the visual sequence. However, a KG can guide inference by searching for possible connections between concepts appearing on images, and thus acquire imaginary concepts.

\subsection{Datasets}
There are no dedicated datasets for K-VIST. On the contrary, relevant literature relies on datasets used for the knowledge-free version of the task, such as \textbf{VIST} \cite{vist}. This dataset contains more than 81k unique photos in around 20k sequences with corresponding textual stories. Textual stories follow a narrative style imposing more high-level inference capabilities compared to literal visual descriptions. This requirement is an extension against the majority of visual description tasks, which do not directly focus on sequential coherence and even abstract meanings. Two extra descriptions are provided per frame in order to bridge literal description with narratives: descriptions of images-in-isolation (DII) and images-in-sequence (DIS).

\subsection{Methods}
\subsubsection{Sequential language models}

\paragraph{External knowledge}
A two-stage structure was proposed in \citet{story1}, consisting of a reasoning and a generation module. The vision-aware commonsense reasoning module is responsible of extracting the most relevant knowledge from an external knowledge base. Objects detected in all images of a sequence are fed in a GRU which provides a semantic and temporal representation. At the same time, candidate ConceptNet entities are fetched based on the detected objects. Attention modules finally select the most relevant ConceptNet candidates, which after passing through a GRU provide the final commonsense representation. 
The knowledge-augumented generation module receives the extracted commonsense knowledge together with the visual information, as well as the previously generated sentences.

A prevalent issue in VIST is the monotonous and repetitive generated stories. This can be attributed to the limited vocabulary of the VIST dataset. In KG-Story \cite{story2}, the first stage (\textit{distill}) gathers words from images using object detection and leverages GRUs for word prediction. Potential relationships between pairs of concepts throughout images are searched on external KGs, and if multiple candidates occur, a scoring function is used to rank their relevancy. This is the \textit{enrich} stage. Finally, the \textit{generate} stage utilizes a Transformer which imposes a repetition penalty to mitigate redundant narration. Further modifications in the default Transformer structure is the introduction of an anaphoric expressions generator to enhance coreferences and usage of pronouns, as well as positional encodings of variable length to enable representing stories of different lengths.

Addressing again the coherence and novelty of generated stories, authors of \cite{story3} propose a three-stage structure corresponding to imagination, reasoning and writing capabilities of humans. The first stage (\textit{imagine}) focuses on the sequential consistence by extracting the visual topic of a frame through the combination of the current visual features and the sentence generated in the previous step. The knowledge part targets the content of narratives and comprises three graph types: a general commonsense KG, a scene graph and an event graph. A GCN applied on each graph selects the most suitable knowledge parts, which are combined to form the second stage (\textit{reason}). Both \textit{imagine} and \textit{reason} outputs are fed to the third stage (\textit{write}), which is responsible for generating the story.

\subsubsection{Transformer-based models}
\paragraph{External knowledge}
Towards informative and more diverse stories, \cite{story4} is the first knowledge-enhanced approach that utilizes a generative transformer to produce the story output. The concept enrichment stage connects concepts present in images with ConceptNet. Then, a graph attention network (GAT) operates on the graph and image features in order to integrate information of the most appropriate candidate concept nodes, which are passed to the next selection module. The concept selection module utilizes two different selection methods: a Sequential Selection Module (SSM) that operates in an encoder-decoder fashion, outputting selected concepts after encoding the embedded candidate concepts; a Maximal Clique Selection Module (MCSM) outputs a maximal clique containing all appropriate concepts for story generation given the concept graph. Finally, the concept to story module uses either an RNN structure or a BART language model, with BART demonstrating more diverse stories while preserving quality.   

\subsection{Evaluation}
\textbf{Language generation metrics} BLUE, ROUGE, METEOR and CIDEr are widely used automatic metrics that evaluate the linguistic quality of generated stories.

\textbf{Diversity of generated stories} is measured via the \textbf{Distinct-n (Dist-n)} score \cite{distn}. This metric indicates the originality of generated text by calculating the frequency of n-grams throughout the whole corpus of generated stories. Higher Dist-n scores represent more diverse stories. \cite{story1, story4}

\textbf{Human Evaluation} is very important for generative tasks, as automatic evaluation metrics cannot assess the full range of linguistic capabilities, especially when it comes to evaluating sequential quality. However, different implementations perform varying human evaluation experiments, which somehow impedes the direct comparison of models. 

In \cite{story1} four aspects are examined: \textit{Fluency} checks the linguistic quality, \textit{relevance} measures the success of textual descriptions in describing visual concepts, \textit{informativeness} measures the diversity of produced stories and \textit{coherence} evaluates the semantic continuity of stories in a sequence. Each aspect receives a score from 1 (worse) to 5 (best) from three evaluators, and their average values serve as the final results. Similarly, in \citet{story3} \textit{relevance}, \textit{coherence} and \textit{informativeness} are regarded, receiving scores from 0 (worse) to 2 (best) from five evaluators. A different human evaluation strategy is followed in \cite{story2}: comparative experiments between VIST models are performed, asking users to rank generated stories from different models either with or without revealing the corresponding images. This is an indirect evaluation of linguistic quality and coherence, when only text is regarded, and also semantic relevance, when corresponding images are provided. The comparative approach is also used in \citet{story4}, with two evaluators declaring their preference (or tie) between two models regarding three aspects: \textit{relevance} and \textit{informativeness} similar to \cite{story1, story3}, together with \textit{logicality} which measures the logical coherence over story sequences. Additionally, \textit{overall} indicates the evaluator's preference in general between the two models.

\section{Knowledge in Image Generation}
\subsection{Datasets}
A variety of datasets have been used in visual generation, which however do not contain some certain demand or sense of knowledge, and are widely used in knowledge-free settings. Datasets used in conditional image synthesis are ImageNet \cite{imagenet}, CIFAR \cite{cifar}, FFHQ \cite{ffhq}, Oxford Flowers \cite{oxfordflowers}, CUB \cite{cub} and many others.

Sequential synthesis (Story Visualization) greatly utilizes \textbf{Pororo-SV} cartoon dataset \cite{pororo-sv}. It contains more than 16k pairs of scenes and dialogs extracted from 20 hours of video,  27,328 fine-grained scene descriptions in natural language provided by human annotators, and 8,913 QA multiple-choice pairs related to the story. In total, 10 main characters appear in the frames. Questions are divided in 11 types: \textit{Action, Person, Abstract, Detail, Method, Reason, Location, Statement, Causality, Yes/No, Time}. \textbf{FlinstonesSV} \cite{flinstones-sv} is also based on cartoon frames. It is composed of 25,184 densely annotated videos, each of which containing 75 frames. The annotations include bounding boxes with labels for characters and items of the frames, as well as segmentation masks.
Another emerging dataset for Story Visualization is \textbf{DiDeMo-SV} \cite{storydalle, didemo-sv}, a dataset based on video captions that contains 10,000 with more than 40,000 temporally localized textual descriptions.

\subsection{Methods}
\subsubsection{Knowledge in Conditional Image Generation (K-cIG)}
\paragraph{Internal knowledge}
Even though GANs have been powerful in synthesizing novel images, they cannot handle combinations of attributes they have not encountered in the training data. Therefore, if the textual condition refers to such unseen combinations, the synthesized image has sacrificed some of the semantics, in order to produce a result that remains within the learned distribution. The insertion of additional knowledge can expand the generated distribution to enhance the consistency on the condition without sacrificing resulting fidelity. This can be translated into two needs regarding a GAN model: the generator should become more flexible, and the discriminator more tolerant. KG-GAN meets those requirements by introducing a second generator, trained on domain knowledge by utilizing a novel knowledge loss. This second generator shares parameters with the original one, which is responsible of synthesizing images conditioned on text. A regression network receives the synthesized images from the seen-image generator and the ones from the knowledge generator, imposing constrains regarding the plausibility of unseen combinations. The semantic vector produced by the knowledge generator is redirected to the seen-image generator to guide generation outside the predefined classes. KG-GAN does not exploit external knowledge sources, but with this simple distribution enhancement it achieves some preliminary zero-shot capabilities. \cite{kg-gan}

\subsubsection{Knowledge in Story Visualization (K-SV)}
\paragraph{External knowledge}
Story Visualization is another task with limited contributions in knowledge-enhanced settings. Structured information from text can be obtained via parse trees which can permit hierarchical encoding of longer phrases. Missing information regarding visual details in text can be filled out with external knowledge from ConceptNet \cite{conceptnet}. Moreover, conceptually similar sentences that are phrased in a different way need to be placed closer in an embedding space, an issue that external knowledge can again effectively resolve. Spatial knowledge is also underrepresented in most sentences, even though scene synthesis needs detailed information of object positions. Dense captioning as a form of self-augmenting knowledge provides detailed positioning information due to the usage of region bounding boxes. The combination of internal spatial and external semantic knowledge is able to better guide sequential synthesis, resolving all involved aspects such as text-image consistency, visual quality and sequential continuity. A Memory-Augmented Recurrent Tree-Transformer
(MARTT) encodes the parse trees for the text, while a Graph Transformer \cite{gtn} embeds the commonsense knowledge. Both embeddings are inserted in the story encoder, which outputs contextualized embeddings for the image generator. The generated images are passed to image and story discriminators, which redirect synthesis based on individual and sequential aspects. Spatial knowledge from dense captioning enforces additional loss functions while training, to provide more explicit information about positions and detailed grounding of characters on the images with respect to their descriptions in the text. \cite{Maharana2021IntegratingVL}

The groundbreaking success of DALL-E \cite{dalle, dalle2} inspired the usage of massive zero-shot transformer-based generative models in Story Visualization; StoryDALL-E \cite{storydalle} achieves generalization of visual synthesis to unseen textual stories, also extending the task to \textit{Story Continuation}: in this case, a source image is included in the conditioning, requesting from the model to continue the visual story in a consistent way. Story Visualization has been a task lacking sufficient datasets, due to the increased effort needed to construct appropriate ones, either manually or automatically. To this end, external unstructured knowledge obtained from the pre-trained DALL-E \cite{dalle} as knowledge base enables even zero-shot sequential synthesis based on input 'story' text.

\subsection{Metrics}
\textbf{Image generation metrics} (Section \ref{sec:imgen_metrics}) such as \textbf{FID} for seen and unseen classes were used in KG-GAN \cite{kg-gan}. FID is also used to evaluate quality of generated frames independently in \cite{Maharana2021IntegratingVL, storydalle}. \textbf{R-precision} indicates quality by measuring the retrieval capabilities of generated frames over ground truth captions comparing to retrieval using the real frames. \cite{Maharana2021IntegratingVL}

\textbf{Classification metrics}, such as \textbf{Character F1 score} measure the quality of generated characters in predicted images. Also, \textbf{frame accuracy} checks the exact match between semantics of the ground truth and generated frames. \cite{Maharana2021IntegratingVL, storydalle}

\textbf{Language metrics} are also relevant: viewing SV frames as a video, captions for generated frames can be produced using video captioning techniques. \textbf{BLEU} scores evaluate the quality of captions as an indirect measure of visual quality, based on the idea that well-designed semantics will be captured in captions better than low quality ones. \cite{Maharana2021IntegratingVL}

\textbf{Human Evaluation} can reveal the human perception over quality, as in most generative tasks. Specifically for SV, evaluators need to assess results over \textit{visual quality}, \textit{consistence} and \textit{relevance} compared to the previous state-of-the-art model on the same task. \cite{Maharana2021IntegratingVL, storydalle}

\section{Multi-task transformers with knowledge}
\subsection{Methods}
Multi-task models can easily be built using multimodal transformer backbones. Instead of utilizing external KGs as in previous methods, many implementations employ self-knowledge exclusively, by obtaining more structured representations from the existing visual and textual data.

\paragraph{External knowledge}
A natural unification of multiple tasks under the same model would incorporate tasks moving in the same direction, such as cross-modal reasoning tasks or cross-modal-retrieval tasks. Indeed, VQA, VCR and VE were unified in Rationale VT transformer \cite{multi2}, a framework that utilizes visual and linguistic clues to generate free-text rationales. Two knowledge sources attempt to provide reasoning information regarding scenes: first, a grounded situation recognizer \cite{grounded_situations} describes activities on scenes with entities involved and draws bounding boxes for entities to visually ground them; second, Visual Commonsense Graphs \cite{visualcomet} are employed to fuse commonsense inferences about events and intents so that a temporal perspective of a scene is also considered. Rationales are generated for VQA-E (visual question answering) \cite{e-vqa}, E-SNLI-VE (visual entailment) \cite{e-SNLI-VE} and VCR (visual commonsense reasoning) \cite{vcr} datasets. Visual recognition of objects is the first step for visual understanding, followed by capturing their in-between relationships utilizing the knowledge provided by the grounded situation recognizer \cite{grounded_situations}. Higher-level cognition is achieved using knowledge from VisualCOMET \cite{visualcomet}, which receives the knowledge stored in Visual Commonsense Graphs to generate commonsense inferences. VisualCOMET is built upon GPT-2 \cite{gpt2}, therefore a unimodal, purely linguistic input can be provided, utilizing object labels, textual question/answers and inferences. Alternatively, GPT-2 can be adapted, resulting in a hybrid implementation: visual features and bounding box coordinates act as visual embeddings, combined with VisualCOMET token embeddings indicating the beginning of \textit{before, after, intent} inferences.

Targeting again reasoning tasks, \cite{multi3} builds on top of LXMERT \cite{xlxmert} to address the knowledge-enhanced versions of the VQA, VCR and VE tasks on the OK-VQA \cite{okvqa}, FVQA \cite{fvqa}, NLVR2 \cite{nlvr2}, SNLI-VE \cite{snli-ve} datasets. External knowledge is provided from ConceptNet \cite{conceptnet} and Wikidata \cite{wikidata}. Knowledge-rich expressions are created by matching embedded knowledge with training sentences from the datasets. Moreover, a training objective targeting the alignment of knowledge embeddings and knowledge-rich expressions encourages learning a global representation structure. Utilizing this objective is proven beneficial during both pre-training and fine-tuning. It is also observed that the introduction of this knowledge-oriented objective smooths the embedding space, which facilitates similarity matching between words.

KB-VLP \cite{multi6} utilizes knowledge embeddings based on Wikidata \cite{wikidata} entities, which are concatenated with the visiolinguistic instances as inputs of a VL transformer. Specifically, entity recognition on text is performed to extract relevant Wikidata entries, which are embedded via Wikipedia2vec to form text-related knowledge embeddings. Object tags extracted from the image are used to obtain image-related knowledge embeddings from relevant Wikidata entities. The input vector consists of 5 components: word embeddings for text, text-related knowledge embeddings, word embeddings sequences for object tags per image, visual features and image-related knowledge embeddings. Two specialized pre-training objectives are used: sentence-level objective substitutes elements from the input vector with other random elements, while token-level objective extends text - image masking to text-related knowledge embedding - image-related knowledge embedding masking. Task specific datasets for KB-VLP are VQA \cite{vqa}, GQA \cite{gqa} and OK-VQA \cite{okvqa} for visual question answering, and NLVR2 \cite{nlvr2} for visual reasoning.

\paragraph{Internal knowledge}
OSCAR \cite{li2020oscar} is one of the models that effortlessly transit from knowledge-free to knowledge-enhanced learning utilizing self-acquired knowledge in its simplest form. Instead of -rather naively- letting the model infer the correct image-text alignments in an exhaustive way, OSCAR facilitates the procedure with the usage of object tags, as intermediaries between text and image instances. This procedure is endorsed from the observation that salient objects in the image will most probably also appear in text. The input to the VL transformer module consists of word tokens, object tag embeddings and visual features. The intermediary object tags form separate semantic spaces, depending on whether they are paired with text or image, yielding two dedicated pre-training objectives. The masked token loss objective views text and tag word representations in the same space, randomly masking each of them and letting reconstruct the missing parts through the visual modality. Conversely, contrastive loss views tags paired with visual features, and randomly replaces the real tag sequence with another one sampled from the dataset, learning to pull apart mismatched tag sequences and bring close together the matching ones. OSCAR succeeds in both understanding tasks, such as cross-modal retrieval (ITR/TIR), visual question answering (on VQA \cite{vqa} and GQA \cite{gqa}) and visual reasoning (on NLVR2 \cite{nlvr2}), as well as in generation tasks, such as image captioning and novel object captioning.

ERNIE-ViL \cite{multi4} leverages structured visual knowledge from scene graphs to bridge detailed semantics across vision and language. Such fine-grained representations are important to differentiate between conceptually similar scenes. Scene graph prediction tasks (object, attribute and relationships prediction) encourage learning those fine-grained differences. Even though not using external knowledge, ERNIE-ViL internally constructs structured knowledge during the cross-modal pre-training. Nevertheless, this self-knowledge is sufficient to boost performance in 5 VL tasks, especially in those that fine-grained associations are required, such as visual referring expressions (VRE). Other tasks benefited from this approach are VCR, VQA and cross-modal retrieval (ITR/TIR).

ROSITA \cite{multi5} extends the self-knowledge idea by employing both cross-modal and intra-modal knowledge at the same time. Given an image-text pair, the first step is to construct intra-modal graphs, i.e an image graph and a text graph. The image graph consists of regions (defined by a pre-trained object detector) as nodes, with IoU scores of paired regions acting as edge weights between those regions. Similarly for the text graph, objects, attributes and relationships are extracted from text to fill the nodes of the text graph, while edge weights are defined by the co-occurrence frequency between pairs of nodes. In both graphs, zero similarity scores between nodes indicate absence of edge. A cross-modal scene graph is derived from the image and text graph by aligning predicted region tags from the image side and words from the text side by comparing their textual semantic similarity. By calculating this similarity score for all possible tag-word pairs, edge weights between cross-modal nodes are defined. Nodes connected via cross-modal edges, named anchor nodes, form sub-graphs which maintain intra-modal and cross-modal edges, as well as two-hop connections that contain paths of cross-modal edges followed by intra-modal ones. ROSITA leverages this enhanced representation to boost three downstream tasks: VQA, VRE and ITR.

\subsection{Evaluation}
\textbf{Human Evaluation} is useful in cases of language generation tasks, such as the rationales generation of \cite{multi2}. In this case, the need for human evaluation arises from the observation that certain discrete rationales, even though not being paraphrases of each other, can be suitable. The following aspects are evaluated: \textit{visual plausibility} referring to how well the generated rationales support the answer (in VQA and VCR) or the entailment (VE) given the image, and \textit{visual fidelity} measuring the appearance of irrelevant information within more plausible generated rationales. By excluding images, \textit{textual Plausibility} evaluates generated rationales based on their support on the answer (in VQA and VCR) or the entailment (VE) exclusively.

\textbf{Classification metrics} such as \textbf{accuracy} serve as the gold standard for non-generative models on cross-modal reasoning tasks. In \cite{multi3}, OK-VQA accuracies per question types are also reported, in order to validate improvements in commonsense-oriented categories attributed to the injection of commonsense knowledge. 

\textbf{Ranking metrics} provide valuable insights in cases when retrieval tasks are performed \cite{multi4, multi5}, where Recall@k for k=1, 5, 10 is reported. 

\textbf{Language metrics} such as \textbf{BLEU} \cite{bleu}, \textbf{CIDEr} \cite{cider}, \textbf{SPICE} are used for language generation tasks, such as image captioning and novel object captioning \cite{li2020oscar}.

\section{The future of knowledge in VL}
\label{sec:future}
\subsection{Explainability and biases}
\label{sec:explain}
Some early works in KVL tasks widely addressed the need for boosting explainability via knowledge graphs. The complex opaque reasoning accompanying many state-of-the-art VL models indeed renders explainability a significant aspect. However, as development of more advanced models emerged, interpretability and fairness became incidental to performance. The pursue of impressive results often promotes models with more vulnerabilities, which have been significantly underexplored. For example, the leading field of NLP has experienced some non-negligible failures, such as producing completely wrong statements based on a false conditioning\footnote{\href{https://openai.com/blog/instruction-following/\#birds-not-real}{Aligning Language Models to Follow Instructions}}. It is unknown how many such vulnerabilities exist in state-of-the-art models, as there is no systematic way to capture and recognize them, nor is there any guarantee they will not occur. Those issues question the trust of humans over black-box models and even make such models susceptible to misuse. Explainability and robustness of VL models are even less explored compared to NLP, leaving lots of space for the development of transparent models and post-hoc explainability methods.
Generally, given the interwoven nature of knowledge graphs and explainability, it is expected that sooner or later research interests will resume towards this direction.

\subsection{Exploitation and integration of more knowledge senses} 
Despite the increasing interest towards knowledge-enhanced multimodal learning, there are some significantly underexplored knowledge aspects that could be leveraged in various tasks. For example, factual knowledge does not have a noticeable presence outside of VQA applications. Named entities and events are only addressed in a couple of works. Temporal knowledge or even hypotheses and counterfactual thinking could reveal new aspects of existing tasks with the potential of interesting implementations.

\subsection{Zero-shot learning}
Previous works regarding zero-shot classification tasks \cite{zsl1, zsl-kg, zsl2, ontozsl} leverage knowledge graphs to fuse feature information from seen to unseen classes. Little work has been done so far towards the more complex task of multimodal zero-shot learning with external knowledge \cite{vqa21}, leaving numerous unexplored directions open for future research. The effortless zero-shot capabilities that Large Language Models (LLMs) present, make them suitable knowledge source candidates. It is expected that more and more knowledge-enhanced works will integrate LLMs within their pipeline, even though this comes at a cost: computational resources and trustworthy generation should be sacrificed in order to harness the rich implicit LLM knowledge, highlighting an inherent trade-off in knowledge source choices.

\subsection{Language Models as Knowledge Bases}
A more natural integration of several knowledge senses can be facilitated via the usage of Language Models as Knowledge Bases (LM-as-KB) \cite{alkhamissi2022review}, a paradigm that has demonstrated some merits towards LM-enhanced approaches, especially in cases where Large Language Models (LLMs) are selected. There is some debate \cite{lymperaiou2023contribution} regarding the risks and challenges of integrating LLMs within the knowledge enhancement pipeline; such issues involve reasoning deficiencies \cite{huang2023reasoning} -especially when  irrelevant content \cite{shi2023large} or misleading inputs \cite{kassner2020negated} are provided-, including the possibility of extensive memorization \cite{Yuan2022CanPL} and confusion between impossible and unlikely events \cite{unlikely}. Nevertheless, the ample interest in LLM reasoning challenges \cite{wei2022emergent, suzgun2022challenging, huang2023reasoning, dziri2023faith, yu2023teaching, pan2023logiclm, morishita2023learning, ho2023large, paul2024making, giadikiaroglou2024puzzle} can shed light and resolve related shortcomings in the near future, rendering LLMs more trustworthy reasoners, and therefore more appropriate knowledge sources.

\subsection{Datasets} 
Even though dedicated knowledge-based datasets have been developed for VQA \cite{vqa5, fvqa, okvqa, vqa6, vqa8, vqa20}, lack of corresponding ventures has been observed in other VL tasks. Such datasets could either incorporate external knowledge from certain knowledge bases in the first place, or be more flexible and dynamically retrieve knowledge to satisfy more challenging inputs. Suitable and high-quality datasets are the first step towards the evolution of the knowledge-enhanced multimodal learning field. Furthermore, knowledge-based datasets could be combined with explanation-oriented ones, such as VQA-E \cite{e-vqa} and e-SNLI-VE \cite{e-SNLI-VE} to address the issues mentioned in Section \ref{sec:explain}.

\subsection{Knowledge-enhanced generative tasks}
The field sitting on the intersection of KGs and generative models has been significantly underexplored, despite the major success both fields have experienced in recent years. Until now, visual knowledge has been attempted in image synthesis, forming the task of image generation from scene graphs and layouts with interesting results and improvements on complex scene synthesis \cite{johnson2018image, pastegan, he2021context}. Also domain knowledge \cite{kg-gan} in GANs has demonstrated some insightful preliminary observations regarding knowledge-guided synthesis of unseen attributes, without the need for massive pre-training. A handful of aforementioned generative multimodal approaches \cite{story1} incorporate knowledge graphs for the tasks of Visual Storytelling and Story Visualization. More compelling results could unfold from the combination of various knowledge graphs with multimodal generative models, enabling conditioning on commonsense, hierarchical, factual knowledge, and also enforcing interpretable insights into the generation process. 

\subsection{The need for multi-task learners}
An abundance of multi-task knowledge-free transformer-based VL models have emerged in recent literature, presenting impressive results on a variety of downstream tasks by utilizing the same pre-trained body each time. On the other hand, knowledge-enhanced VL models usually target a single task, and only a few knowledge-enhanced multi-task models \cite{multi2, multi3, multi4, multi5, multi6} have been developed. Even in the case of multi-task transformers, almost half of them utilize only self-knowledge without exploiting the benefits of additional external sources. At the same time, the harder venture of integrating external knowledge limits the range of tasks that multi-task models target. Specifically, current implementations have covered only reasoning tasks. Therefore, as a first future direction, knowledge-enhanced retrieval tasks can also be attempted. Going one step further, some VL tasks addressed in multi-task knowledge-free VL transformers have never been explored in literature; hence, unified multi-task architectures could possibly explore their knowledge-enhanced capabilities, without the need for developing individual non-reusable approaches. In any case, multi-task knowledge-enhanced models would unlock the full potential of the contributions of knowledge, with competitive architectures pushing the state-of-the-art even further.

\section{Conclusion}
Introducing external knowledge in multimodal learning has demonstrated promising research directions, targeting performance, explainability and extendability of existing tasks. In this survey paper, we analyzed the meeting point of visiolinguistic (VL) representation learning and knowledge-enhanced learning, focusing on the contribution of existing knowledge graphs and unstructured knowledge sources, such as language models. The presented taxonomy of knowledge-enhanced datasets, tasks and models provides one of the first attempts towards structuring the field of knowledge-enhanced VL learning, with the aim to guide future research and address prospects and challenges of this upcoming field.

\bmhead{Acknowledgments}
The research work was supported by the Hellenic Foundation for Research and Innovation (HFRI) under the 3rd Call for HFRI PhD Fellowships (Fellowship Number 5537).

\bibliography{sn-article}

\end{document}